\documentclass[9pt, table, x11names]{article} 

\usepackage{branding}
\usepackage{authblk}
\usepackage{caption}
\usepackage{subcaption}
\usepackage{graphicx}
\usepackage{multicol}
\usepackage{geometry}
\usepackage{rotating}
\usepackage{comment}   % remove before completion?  only while editing.

\usepackage[parfill]{parskip}

\begin{document}

%%\togglecolumns %no need to uncomment. Suppresses two columns for report but leave two columns for preprint

%%\input{branding/titlepage}
% \input{branding/docrecord} %uncomment if applicable
%% \input{branding/contents} 

\title{\textbf{Applying Psychometrics to Large Language Model Simulated Populations: Recreating the HEXACO Personality Inventory Experiment with Generative Agents}}

\date{\vspace{0ex}}

\author[1]{Sarah Mercer\thanks{smercer@turing.ac.uk}}
\author[1]{Daniel P. Martin}
\author[1]{Phil Swatton}
\affil[1]{The Alan Turing Institute}

\maketitle
\begin{abstract}
\noindent Generative agents powered by Large Language Models demonstrate human-like characteristics through sophisticated natural language interactions.  Their ability to assume roles and personalities based on predefined character biographies has positioned them as cost-effective substitutes for human participants in social science research.  This paper explores the validity of such persona-based agents in representing human populations; we recreate the HEXACO personality inventory experiment by surveying 310 GPT-4 powered agents, conducting factor analysis on their responses, and comparing these results to the original findings presented by Ashton, Lee, \& Goldberg in 2004.  Our results found 1) a coherent and reliable personality structure was recoverable from the agents' responses demonstrating partial alignment to the HEXACO framework. 2) the derived personality dimensions were consistent and reliable within GPT-4, when coupled with a sufficiently curated population, and 3) cross-model analysis revealed variability in personality profiling, suggesting model-specific biases and limitations.  We discuss the practical considerations and challenges encountered during the experiment. This study contributes to the ongoing discourse on the potential benefits and limitations of using generative agents in social science research and provides useful guidance on designing consistent and representative agent personas to maximise coverage and representation of human personality traits.

\end{abstract}
\vspace{10mm} 

%title, authors and abstract

\renewcommand{\thefootnote}{\roman{footnote}} 
\def\UrlBreaks{\do\/\do-} %hack to cope with long urls

\section{Introduction}

One of the primary goals of Artificial Intelligence is to develop systems that can replicate human behaviours and interactions.  The field of Natural Language Processing (NLP) plays a crucial role in enabling AI systems to understand and generate human language.  Recent advancements in Generative AI, specifically Large Language Models (LLMs), have enhanced these capabilities, raising significant interest in such systems’ ability to mimic human cognitive, emotional, and social behaviours and resulted in an increased use of generative agents for social simulations~\cite{gurcan2024, hua2024, park2023simulacra, horton2023, guo2024survey, aws_blog}, within computational social science~\cite{ziems2024, lazer2020} and the emerging field of machine psychology~\cite{hagendorff2024, li2024quantifyingaipsychology, pellert2024}.

In parallel with the developments in social simulations, a separate body of research examines the extent to which LLMs can be used to simulate human participants in experiments, opinion polls, and surveys in social scientific research~\cite{petrov2024, bail2024, argyle2023, simmons2023}. There are several perceived benefits, such as reduced costs, collecting hard-to-acquire data, or simply reduced administrative burden.

What makes generative agents appropriate for both types of simulation is the data upon which the encompassed LLMs have been trained, the huge amount of human generated text, in the form of books, website articles and social media content, which inherently describes a diverse set of human expressions, interactions, and nuances of language use commonly exposed by humans.  As such, generative agents can mimic (to varying degrees, and not without some contention~\cite{bender2021, marcus2020}) linguistic styles, knowledge, and cognitive processes present in the training data, which can appear to be humanlike.  However, as this mimicry is fundamentally computational (next word prediction) it raises the question of how successfully these agents emulate human behaviour (i.e. how valid are they beyond surface level responses).

\subsection{Related Work}

To quantify the inherent qualities and attributes of LLMs and generative agents, AI researchers have drawn on the field of psychology. Several studies have explored the psychological aspects of LLMs, investigating whether the cognitive and reasoning abilities these models appear to display align closely with those observed in humans. Notably, Hagendorff \textit{et al}., used several behaviour tests; the Cognitive Reflection Test (CRT)~\cite{shane2005} and semantic illusions~\cite{ericson1981} designed to investigate intuitive decision-making in humans (e.g. fast ‘v’ slow thinking), concluding that ‘investigating LLMs with methods from psychology has the potential to reveal otherwise unknown emergent traits’~\cite{hagendorff2023}.

Macmillan-Scott and Musolesi~\cite{macmillan2924} used “cognitive illusion” tests (brain teasers) originally designed by Kahneman and Tversky~\cite{kahneman1972} to illustrate cognitive biases and heuristics in human reasoning, to evaluate the rationality of LLMs.  They noted that given the fact that responses varied for the same prompt and model, a slightly different approach to evaluating LLMs would be required.  They also observed that the models’ incorrect responses were ‘incorrect in ways different to human subjects’.  

In addition to cognitive testing, psychometric tests have also become popular for measuring the personalities demonstrated by Generative AI.  In 2022, Miotto \textit{et al.}~\cite{miotto2022} employed HEXACO-60~\cite{ashton2009} on OpenAI's Davinci, noting that its results were similar to human samples when provided with a history of previous responses.  There have been many similar studies since then, the results of which have varied due to differences in choice of psychometric test, prompting strategy, and  degree of academic rigour.

Even as the approaches mature, researchers are still reporting mixed results.  Safdari \textit{et al.}~\cite{safdari2023} used two measures; IPIP-NEO~\cite{goldberg1999} and BFI~\cite{john1999}, and a principled methodology to explore how a model's size and training procedures affected simulated personalities traits, finding that fine-tuning increased the stability of personality structures. Gupta \textit{et al.}~\cite{gupta2024} argued against that conclusion, suggesting that validity tests must also be rooted in LLM understanding and not human-based best practice. Using IPIP-300~\cite{johnson2014}, they concluded that due to LLMs' unreliability at answering multiple choice questions (MCQs), any personality scores derived from MCQs would also be unreliable.  Huang \textit{et al.}~\cite{huang2024-reliability} also used BFI to test the degree to which varying the format of the assessment questions affected the resulting personality scores, showing that GPT-3.5-Turbo, GPT-4 and Gemini-Pro were able to generate stable responses across diverse settings. All three models exhibited tendencies towards specific personality traits. Muhua Huang \textit{et al.}~\cite{huang2024} concluded that although agents could be assigned personas that performed similarly to humans in psychometric tests (BFI), these assigned traits did not translate into personality-consistent behaviours.  In particularly, the agents' risk-taking behaviours aligned with human patterns, but ethical dilemma decisions did not. de Winter \textit{et al.}~\cite{dewinter2024} generated 2,000 personas, evaluated them using the BFI-10~\cite{rammstedt2007}, and found their results aligned closely with human baselines.  However, when evaluating a second persona set, they found that the correlation matrix among personality constructs was affected by persona set.  

Psychometric tests such as BFI-10 and HEXACO-60 are short-form tests developed for practical human assessments and are secondary interpretations of broader personality structure theories~\cite{rammstedt2007, ashton2007, ashton2009}.  Applying these measures to LLMs compounds the conceptual limitations, as they assume a stable identity and the ability for introspection, neither of which truly apply to LLMs.  Consequently, any results from such psychometric testing of LLMs should be interpreted with caution, as outcomes may reflect superficial patterns rather than meaningful personality attributes, leading researchers to draw misleading conclusions about the capabilities and validation of generative agents as accurate proxies for human participants in social science.

\subsection{Our Contribution}

To explore the validity of applying psychometric tests to LLMs, this research replicates the foundational experiment that established the six-factor structure underlying the HEXACO personality inventory~\cite{ashton2004}.  Originally conducted in 2004 by Ashton, Lee and Goldberg, the lexical analysis of human responses (\textit{N}=310) revealed that the structure of personality traits in the English language was similar to findings from other languages, suggesting that personality traits may be universal.  To our knowledge, this work represents the first attempt to specifically recreate the foundational HEXACO lexical study using LLMs.

Given that LLMs do not possess personalities in the human sense, acknowledging that human personality dimensions are themselves statistically derived constructs rather than direct representations of neural structures, the focus shifts to understanding how consistently and effectively these models can simulate human-like behaviours.  Which is crucial for ensuring reliability and validity in social science research of using agents.  Recent literature suggests that while LLMs can replicate certain aspects of personality, the extent of their mimicry depends on the model and context~\cite{gupta2024, hilliard2024}.  Achieving realistic human-like behaviour in generative agents can enhance user experience, build trust, and inform the refinement of AI models. At the same time, highly realistic behaviour also introduces ethical concerns such as manipulation or deception, which must be mitigated through evaluation and design.

This research aims to assess whether agent-generated responses produce comparable personality structures to those identified in human-based lexical studies.  By closely recreating Ashton \textit{et al.}'s original methodology, but with agents replacing human participants, this study provides unique insights into both the capabilities and limitations of agent-based personality assessments.  Findings from this study will help establish boundaries for the use of agents in social science, contributing directly to debates in both social science and machine psychology.

\section{HEXACO Experiment}

\subsection{Persona Creation}

Previous research has demonstrated multiple ways to prompt an LLM to adopt a persona; for example, by directly specifying the personality characteristic and traits an agent is to adhere to~\cite{jiang2024}, by providing descriptive narratives (often machine generated) to utilise the stereotypical/caricature information encoded within the LLM~\cite{park2023simulacra, mercer, dewinter2024}, or by selecting personality descriptors from human-generated data sets~\cite{safdari2023, miotto2022, serapiogarcía2024, zhang2018}.  Huang \textit{et al.}~\cite{huang2024} used completed 60-item BFI-2~\cite{bfi2} profiles from real participants as personas, showing that such profiles could be derived from publicly available personality statistics (means, standard deviations, correlations), and that these simulated profiles yielded very similar results to the human-based prompts.  Argyle \textit{et al.}~\cite{argyle2023} combined several approaches - where a narrative is formed from a structured set of attributes, with additional information harvested from known datasets.

For this study, we followed the approach taken in the Smallville simulacra~\cite{park2023simulacra}. GPT-4 was used to generate structured biographies consisting of key attributes: name, age, occupation, hobbies/interests and personality facts. The structured format ensures personas are systematically comparable whilst still allowing for emergent biases and associations inherent in the LLM to surface.  By prompting the model to generate full biographies rather than just isolated personality descriptors, we can examine how different attributes (e.g. occupation, interests) interact and the extent to which the LLM-generated personas reflect stereotypical or biased associations. The prompts used to generate the biographies, and some example biographies can be found in Appendix \ref{app_prompts}

\subsection{Population Curation}

\begin{figure}[!h]
\centering
  \includegraphics[width=0.8\textwidth]{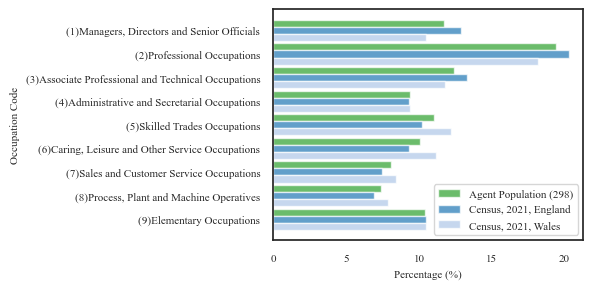}
  \caption{Agent population (PopCensus) broken down by top-level SOC2020 Classifications against 2021 Census for England and Wales.}
  \label{fig:occupations}
\end{figure}

To curate a representative agent population, we aligned the agent occupations to the results of the 2021 Census for England and Wales~\cite{ONS_2022}, across the Standard Occupational Classification (SOC) major groupings~\cite{ONS_SOC}. We called this population `PopCensus'.  We prompted GPT-4 to complete a biography for a given occupation, resulting in a population distribution as shown in Figure \ref{fig:occupations}.

The mean agent age was 39.13 years (\textit{SD}=9.51) ranging from 18 to 60 years\footnote{Specifying the age range of the agent personas generated to between 16 and 60 was to encourage the LLM to focus on the occupation, upon which, many of the personality trait assumptions would be formed.  To fit the 2021 Census, retired and unemployed agents were created, however the upper limit on age was not accordingly adjusted to accurately reflect this additional population.  As such the population is not aligned with the age profile within the 2021 census, just the working status/occupational data.}.  Although the LLM was not explicitly required to specify a gender in the biography; by examining responses we can determine the population contains 200 male and 102 female biographies\footnotemark.  Across the 310 agents, there were 272 unique occupations.
\footnotetext{The original HEXACO study comprised 200 women and 110 men from the US and Australia.}

\subsection{Methodology}

We followed the procedure described in the original paper~\cite{ashton2004}. 310 agents were prompted to self-rate on a set of 1,710 adjectives\footnotemark{} using a 9-point response accurate/inaccurate scale.  The prompt used to conduct the survey required the agents to provide an explanation of their self-rating for each adjective to improve the accuracy of the generated response~\cite{white2023, jung2022}, the prompts used are presented in Appendix \ref{app_prompts}.  The responses were collated and the ipsatisation process (standardising responses between and within individuals), as detailed in the original study, was applied to account for differences in how participants used the response scale. 
\footnotetext{The original set of 1,710 adjectives was undiscoverable, this set was obtained from Harvard-Dataverse~\cite{condon2021}, however there were some differences between the set we used and the original, as evidenced by the lack of ‘gentle hearted’ contained within the top loaded terms in the original paper but not found in the set used on this study.}
We used OpenAI’s GPT-4 (via Azure)~\cite{gpt4azure} for the duration of this experiment\footnote{Temperature=0.7}.

During data collection, Azure's content filter withheld responses from 225 agents when asked about the adjective \textit{niggardly}\footnotemark{}, and therefore we excluded this adjective from subsequent analyses to ensure consistency.  Additionally, GPT-4 failed to return answers for other adjectives on 66 occasions (approximately 0.012\% of total responses). There were seven adjectives upon which all the agents returned the same response: \textit{adulterous}, \textit{flammable}, \textit{homicidal}, \textit{illiterate}, \textit{lewd}, \textit{murderous}, and \textit{sadistic}.  These adjectives were excluded from the remainder of the experiment as they lacked variability. Most of these terms are associated with a Dark Triad personality trait, with the exclusion of \textit{illiterate}.  It is likely that this is a result of the models' safety training and/or behaviour alignment, where models are encouraged to avoid being inappropriate or causing harm.
\footnotetext{Niggardly means ‘slight in amount, quality, or effort’ as defined in the Cambridge Advanced Learner’s Dictionary \& Thesaurus, however, as discussed on the Merriam Webster site, although the words niggard and niggardly are etymologically unrelated to a highly offensive racial slur, the resemblance means both terms are often taken to be offensive.}

\subsubsection{Principal Component Analysis}

\begin{figure}[!h]
  \centering
    \includegraphics[]{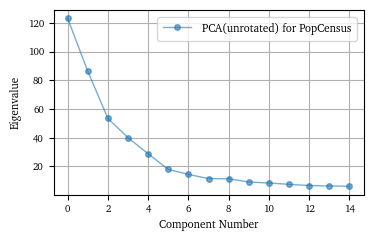}
    \caption{Scree Plot of (unrotated) Eigenvalues for agent responses (PopCensus).}
    \label{fig:scree_plot}
\end{figure}

Eigenvalues were used to identify the optimal number of factors for factor analysis, a scree plot of the (unrotated) eigenvalues is shown in Figure \ref{fig:scree_plot}.  From this plot it appears the 5-factor solution should be studied further.  However, to allow for comparison to the original study we will also explore the 6-factor solution.

The original study used varimax rotation, whilst stating that results based on oblique rotations (e.g. promax) were very similar. For this study we have used promax rotation~\cite{hendrickson}, which allows factors to be correlated.

\subsubsection{Criteria of Successful Simulation}

Cross-linguistic research has consistently identified the first four of the Big 5 factors~\cite{saucier1996} – Extraversion, Agreeableness, Conscientiousness and Emotional Stability (Neuroticism), while the fifth factor intellect-imagination (Openness to Experience) is sometimes replaced by an honesty-related dimension, as observed in Hungarian and Italian studies~\cite{ashton2004}. We therefore take successful simulation of these four core dimensions as a minimum criterion for success in our agent-based approach. We therefore do not consider only standard criteria for selecting a number of dimensions from PCA, but also whether we are better able to improve simulation success.

\subsubsection{Internal Consistency and Cross-validation Measures}

To evaluate the reliability of the dimensions extracted via PCA, Cronbach's alpha~\cite{jim} was calculated for each recovered factor, indicating the internal consistency among adjectives measuring the same underlying trait.  Coverage of adjectives from the original HEXACO study was assessed using a weighted Jaccard similarity metric, accounting for the directionality of factor loadings.  This metric provided a means of cross-validation, directly comparing agent-generated adjective ratings and the original human-derived ratings. \footnote{To calculate the weighted Jaccard similarity, a reduced set of loadings was used to mirror the limited set of loadings presented in the original paper.  The top 30 (absolute) values were selected.}

\subsection{Findings}
\subsubsection{5-Factor solution}

The top loaded terms for the 5-Factor solution are presented in Appendix \ref{app_5_results}.  Based on these adjectives and their directional loadings, we labelled the recovered factors as follows: \textbf{Introversion}, \textbf{Assertiveness}, \textbf{Dishonesty}, \textbf{Unconventionality}, and \textbf{Provincial}.  Variance explained by these factors ranged from 4.84\% to 2\%, with a cumulative explained variance of 19.54\%, slightly higher than the original study’s 4.34\% to 2.37\% (cumulated = 18.56\%).

Cronbach’s alpha reliability scores for these factors were 0.95, -0.56, 0.94, 0.15, -0.39 respectively.  The high alpha values for factors 1 and 3 indicate strong internal consistency among their terms.  However, negative or low alpha values for the other factors highlight issues likely reflecting problematic item groupings.

The weighted Jaccard similarity scores, assessing the coverage compared to the HEXACO dimensions, identified the strongest alignments for each recovered factor as: 1 - Extraversion (0.045), 2 - Agreeableness (0.039), 3 - Honesty-Humility (0.081), 4 - Openness (0.032), and 5 - Openness (0.041), all results are presented in Figure \ref{app_jaccard}.\ref{fig:jaccard_5}.  The alignment with Openness across multiple factors likely reflects the differentiation of facets within this broad dimension, as some adjectives in factor 4 relate to the unconventionality facet of Openness, and some terms in factor 5 relate to the intellectual aspect.

In summary, factor 3 (Dishonesty) demonstrated strong internal consistency and strong alignment with HEXACO's Honesty-Humility dimension.  Factor 1 (Introversion) had strong internal consistency and a reasonable overlap with HEXACO's Extraversion, whereas factor 2 (Assertiveness) had very poor internal consistency but overlapped with Agreeableness fairly well. Factor 5 (Provincial) showed very poor internal consistency despite moderate overlap with terms from Openness, and factor 4 (Unconventionality) exhibited limited cohesiveness and did not clearly align with a specific HEXACO dimension.  Given these limitations, particularly the suboptimal cohesiveness indicated by low or negative alpha values, this solution fails to adequately provide evidence for a 5-factor personality structure in the agent responses. Next, we explore the 6-factor solution, as increasing the factors can often improve reliability.

\subsubsection{6-Factor solution}

The top loaded terms for this solution are presented in Appendix \ref{app_6_results}. Explained variance ranged from 4.24\% to 1.96\%, with cumulated variance at 19.71\%, slightly lower than the original HEXACO study (4.26\% to 1.97\%, cumulative = 20.16\%).  Notably, the top-loaded terms from the agent solution had substantially higher weightings (up to 0.91) compared to the original study (maximum 0.66), suggesting that the agents' responses were more consistent than the original study's human participants, implying a limitation on the ability of agents to simulate human responses.

Cronbach’s alpha values for the factors were 0.94, 0.97, -0.2, -0.47, 0.91 and -0.42, respectively.  This indicates excellent internal consistency for factors 1, 2 and 5, while factors 3, 4 and 6 demonstrated substantial internal consistency issues, as indicated by negative alpha values.

The highest weighted Jaccard similarity alignments to HEXACO dimensions respectively were 1 - Honesty-Humility (0.081), 2 - Extraversion (0.081), 3 - Openness (0.032), 4 - Extraversion (0.015), 5 - Agreeableness (0.15), and 6 - Openness (0.042). All results are presented in Figure \ref{app_jaccard}.\ref{fig:jaccard_6}.  

In summary, factors 1, 2 and 5 had strong internal consistency, cohesive adjective groupings, and aligned well with the original study’s Honesty-Humility, Extraversion and Agreeableness dimensions, albeit primarily representing their negative poles.  As such we labelled them \textbf{Dishonesty}, \textbf{Introversion} and \textbf{Disagreeableness}.  In contrast, factors 3 (\textbf{Unconventionality}), 4 (\textbf{Dominance}), and 6 (\textbf{Provincial}) exhibited weaker internal consistency, less coherent adjective groupings, and ambiguous alignments with HEXACO dimensions.

Overall, the 6-factor solution demonstrates that the agent survey responses possess sufficient complexity and internal consistency to yield interpretable factors.  However, the recovered personality structure notably differs from the human equivalent.

\subsubsection{Beyond Six Factors}

The 6-factor solution performed better than the 5-factor solution, but still only recovered two of the core dimensions; Extraversion and Agreeableness-related factors, and both failed to satisfactorily recover dimensions thematically aligned to Emotionality or Conscientiousness, suggesting a limitation in agent-based systems to replicate the complete human-based personality structure.

The 7-factor solution did recover dimensions that aligned with Emotionality, and (the absence of) Conscientiousness.  Alongside factors describing the absence of Extraversion, Agreeableness, Honesty-Humility and one which described the absence of Intellect, a facet of the Openness factor.  Meaning we recovered, although weighted differently, the four core dimensions.  In the original study, Ashton \textit{et al.}, briefly discussed their additional seventh factor interpreted as Religiosity.  However, the seventh factor recovered from the agents' responses contained terms describing a \textit{boldness} and \textit{adventurous} `v' \textit{soft-spoken} and \textit{leisurely} dimension.  The emergence of this distinct factor suggests that while agent-generated personality structures share similarities with human-derived models, they may also contain unique latent dimensions, potentially reflecting different underlying semantic associations encoded in the language model.

\begin{figure}[!h]
\centering
  \includegraphics[]{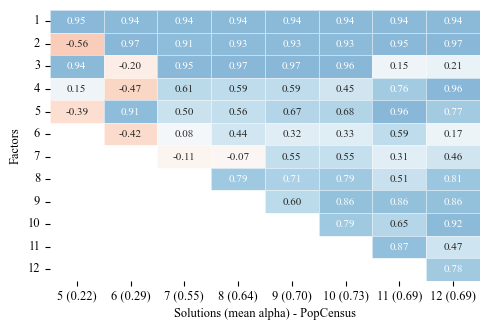}
  \caption{Cronbach’s alpha for each factor recovered across all solutions. Numbers in brackets on x-axis are average alpha value for solution.}
  \label{fig:alphas}
\end{figure}

From this we can determine that although similar in nature, the factors recovered from the agent responses do not broadly reflect the human personality structure. However, the recovered factors are not poor enough to say that no personality structure is recoverable.  To identify more robust solutions, we examined the Cronbach alpha values for all solutions up to a 12-factor solution. The results are presented in Figure \ref{fig:alphas}, showing that the 10-factor solution has the highest average Cronbach's alpha score of all solutions.  We consider this solution to be the most reliable approximation of a personality structure derived from the agents' responses.

\subsubsection{10-Factor solution}

The top-loaded adjectives for the 10-Factor solution can be found in Appendix \ref{app_10_results}.  Explained variance across factors ranged from 4.00\% to 0.92\%, with a cumulative explained variance of 22.51\%, comparable to the original study's variance of 20.16\%.  Cronbach's alpha values for the ten factors are: 0.94, 0.93, 0.96, 0.45, 0.68, 0.33, 0.55, 0.79, 0.86, and 0.79, averaging at 0.73.

Factor 1 strongly reflected terms related to \textbf{Dishonesty}, with the highest loaded terms \textit{sly} (0.94), \textit{sneaky} (0.93) and \textit{deceptive} (0.93).  Among the top 30 terms, only five adjectives loaded in the opposite direction. Factor 2 captured \textbf{Disagreeableness}, with leading adjectives \textit{sharp-tongued} (0.74), \textit{abrasive} (0.69) and \textit{forbearing} (0.67) and only 4 terms weighted towards agreeableness.  Factor 3 consistently captured traits of \textbf{Introversion}, led by adjectives \textit{uncommunicative} (0.72), \textit{aloof} (0.71), \textit{untalkative} (0.71).  Factor 4 contained terms related to \textbf{Unconscientiousness} including \textit{overneat} (-0.67), \textit{overconscientious} (-0.6) and messy (0.6).  

Factor 5’s top-loaded terms include \textit{unadventurous}, \textit{fierce}, \textit{hectic}, \textit{audacious}, and \textit{lion-hearted}, describing a heroic verses sedate dimension – \textbf{Unheroic}. Factor 6’s top-loaded terms were \textit{unbookish} (0.71), \textit{unscholarly} (0.70), \textit{unliterary} (0.67), suggesting a factor related to absence of intellect - \textbf{Unscholarly}, which is one of the facets of Openness.  

Factors 7, 8 and 9 all contain terms related to Emotionality.  Factor 7 focusses on the \textbf{Gendered Emotionality} the top-loaded terms are \textit{womanly} (1.07), \textit{gentlemanlike} (-0.96), \textit{feminine} (0.90) and \textit{masculine} (-0.89).  Factor 8 contains terms reflecting an absence of sentimentality (\textbf{Unsentimentality}) such as \textit{tender-hearted} (-0.36), \textit{poetic} (-0.33), \textit{spiritual} (-0.34), \textit{altruistic} (-0.29), and \textit{dispassionate} (0.28).  Factor 9 includes terms related to the absence of anxiety and sensitivity (\textbf{Insensitivity}) including adjectives: \textit{fretful}, \textit{worrying}, \textit{anxious}, \textit{tense}, and \textit{think-skinned}.  

Factor 10 contained adjectives which include \textit{chic}, \textit{elegant}, \textit{sophisticated}, \textit{showy}, \textit{cosmopolitan}, \textit{artistic}, \textit{untheatrical}, and \textit{inelegant}, describing a lack of imagination-related traits (\textbf{Unartistic}).

\begin{figure}[!h]
  \centering
  \begin{minipage}{0.49\textwidth}
    \includegraphics[width=\textwidth]{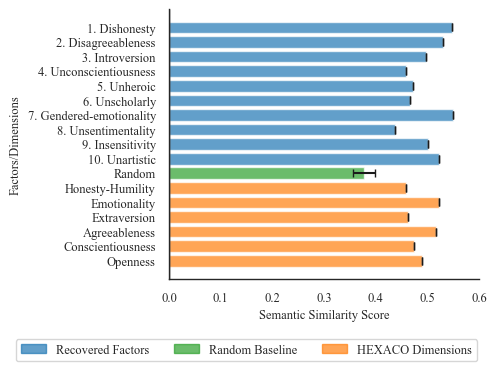}
    \caption{Average Symmetric Semantic Similarity of terms within each factor, across agent's 10-factor solution, and original HEXACO top-loaded terms.}
    \label{fig:average_similarity}
  \end{minipage}
  \hfill
  \begin{minipage}{0.49\textwidth}
    \includegraphics[width=\textwidth]{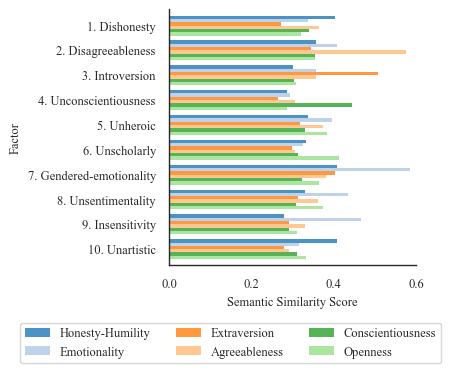}
    \caption{Symmetric Semantic Similarity for 10-factor solution against original study’s HEXACO top loaded terms (using FastText).\newline}
    \label{fig:semantic_sims}
  \end{minipage}
\end{figure}

To explore the cohesiveness of the factors we used Symmetric Semantic Similarity\footnotemark{} to measure how closely related terms within each factor are to each other, see Figure \ref{fig:average_similarity}.  The average semantic similarity across all agent-derived factors was 0.498 (\textit{SD}=0.039), closely comparable to the original HEXACO top-loaded terms at 0.497 (\textit{SD}=0.03).  For additional comparison, we calculated a baseline similarity by randomly selecting set of 25 adjectives across ten iterations, yielding a significantly lower average of 0.363 (\textit{SD}=0.026).  The lower baseline reinforces that the recovered factors represent meaningful and stable clusters akin to those in the original study.
\footnotetext{Measuring cosine similarities between word embeddings using FastText: https://ai.meta.com/tools/fasttext/}

The weighted Jaccard similarity for this solution is presented in Figure \ref{app_jaccard}.\ref{fig:jaccard_10}. Highest similarity values respectively are 1 - Honesty (0.081), 2 - Agreeableness (0.131), 3 - Extraversion (0.083), 4 - Conscientiousness (0.065), 5 - Emotionality (0.024), 6 - Openness (0.047), 7 - Emotionality (0.166), 8 - Emotionality (0.048), 9 - Emotionality (0.089), 10 - Honesty (0.026).  

We also examined semantic alignment between the agent-derived factors and the HEXACO dimensions\footnotemark{} through Symmetric Semantic Similarity, as shown in Figure \ref{fig:semantic_sims}.  This analysis confirmed strong alignment for eight factors: 1-Honesty-Humility, 2-Agreeableness, 3-Extraversion, 4-Conscientiousness, 6-Openness, and factors 7, 8 and 9 collectively aligned with Emotionality, although each captured a distinct facet (gendered-related traits, sentimentality/spirituality, and anxiety/sensitivity).  Factors 5 and 10 did not clearly align with any specific HEXACO dimension, highlighting the emergence of novel semantic groupings unique to the generative agents' responses. 
\footnotetext{Of the top loaded terms as described in the Ashton's paper, only the term ‘gentle-hearted’ from the original study’s Agreeableness factor was not present in the agent’s adjective set.}

In summary, factors were recovered that resemble HEXACO’s six dimensions, with two additional factors 5 (Unheroic), and 10 (Unartistic), showing that the agents have additional personality dimensions related to their outputs.  In the original study, Openness was an umbrella term used to cover intellect, imagination and unconventionality.  However, in the agent-derived solution, these concepts are partitioned into other factors; the intellect aspects are recovered in the 6 (Unscholarly) factor, terms related to imagination such as \textit{untheatrical}, and \textit{artistic} were recovered as part of factor 10 (Unartistic), with terms related to defiance versus submissiveness, such as \textit{untamable} versus \textit{submissive}, being found in the 5 (Unheroic) factor. This suggests that agents differentiate these aspects into separate semantic clusters.

\subsection{Reliability and Validity}

Convergent validity is the degree to which two measures, which theoretically should be related, are actually related, i.e. if the personality structure recovered from the lexical analysis is reliable and robust, each agent's personality should be recoverable via another measure of personality. To explore convergent validity, we tested our agent population (PopCensus) using the HEXACO-PI-R 100 questionnaire, comparing the results against scores derived from lexical analysis, where each agent received a score based on their responses, calculated across each factor.

To evaluate the convergent validity between the lexical analysis and the survey results, Pearson correlation coefficients were determined for each of the six HEXACO dimensions\footnotemark{}.  The analysis revealed statistically significant correlations for most dimensions; Honesty-Humility (\textit{r}=-0.814, \textit{p}=0), Extraversion (\textit{r}=-0.833, \textit{p}=0), Agreeableness (\textit{r}=-0.873, \textit{p}=0), and Conscientiousness (\textit{r}=-0.934, \textit{p}=0).  Emotionality (\textit{r}=0.686, \textit{p}=0) and Openness (\textit{r}=-0.447, \textit{p}=0) show comparatively weaker correlations, suggesting these lexical measures are somewhat less representative or precise; this is unsurprising given terms related to these dimensions were recovered across multiple more-specific factors, these values reflect the correlation between the HEXACO dimensions and the most semantically similar factor for each, in this case `Gendered-emotionality' and `Unscholarly', respectively.
\footnotetext{Of the 10 recovered factors, we chose the six that had the highest Weighted Jaccard Similarity to the original HEXACO dimensions: 1-Dishonesty, 7-Gendered emotionality, 3-Introversion, 2-Disagreeableness, 4-Unconscientiousness, and 6-Unscholarly.}

\begin{figure}[!h]
\centering
  \includegraphics[width=0.6\textwidth]{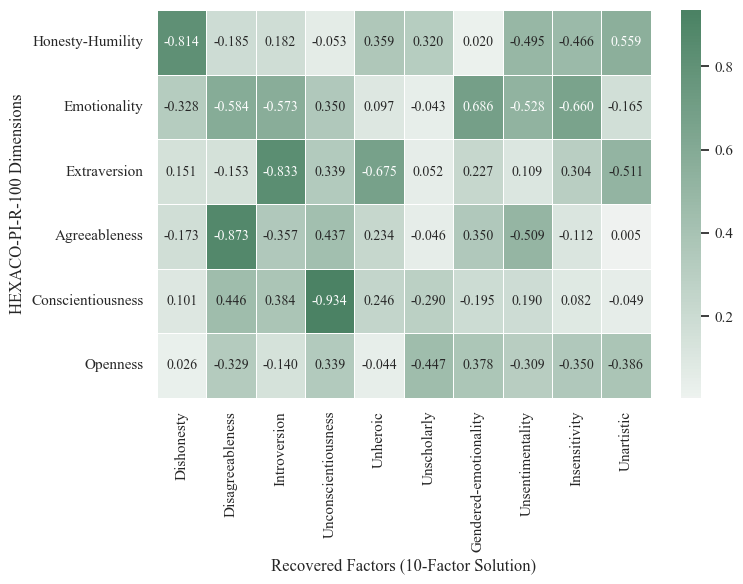}
  \caption{Pearson's correlation coefficients between lexical factors and HEXACO-PI-R 100 results, using GPT-4.}
  \label{fig:correls}
\end{figure}

To further confirm the reliability of the recovered structure, we use discriminant validity to show that measures which should not capture the same trait, are not well correlated.  Figure \ref{fig:correls} shows the mapping between the lexical factors and the survey results.

These findings indicate that personality dimensions inferred from the lexical analysis align strongly with those measured by the HEXACO-PI-R 100 questionnaire, indicating robust convergence between these two methods.  Given that both assessments were conducted using the same underlying LLM, it suggests notable internal consistency within the simulated personality profiles of the agents.  This implies the LLM can produce stable, personality-like dimensions across different measurement approaches.

\begin{figure}[!h]
  \centering
  \begin{minipage}[b]{0.45\textwidth}
    \includegraphics[width=\textwidth]{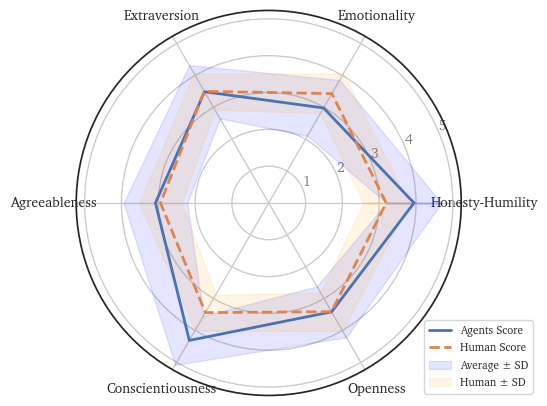}
    \caption{HEXACO-PI-R 100 survey results for PopCensus using GPT-4.}
    \label{fig:pir_gpt4}
  \end{minipage}
  \hfill
  \begin{minipage}[b]{0.45\textwidth}
  \includegraphics[width=\textwidth]{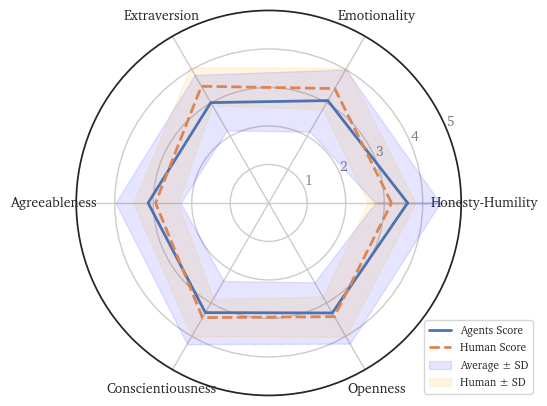}
  \caption{HEXACO-PI-R 100 survey results for PopCensus using Llama 3.2.}
  \label{fig:pir_llama}
  \end{minipage}
\end{figure}

\begin{figure}[!h]
  \centering
  \begin{minipage}[b]{0.45\textwidth}
    \includegraphics[width=\textwidth]{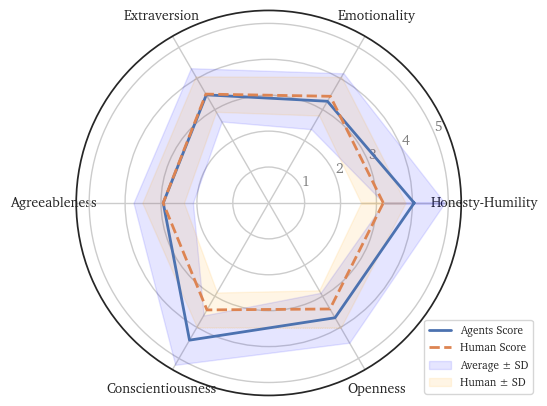}
    \caption{HEXACO-PI-R 100 survey results for PopCensus using Claude 3.7 Sonnet.}
    \label{fig:pir_sonnet}
  \end{minipage}
  \hfill
  \begin{minipage}[b]{0.45\textwidth}
  \includegraphics[width=\textwidth]{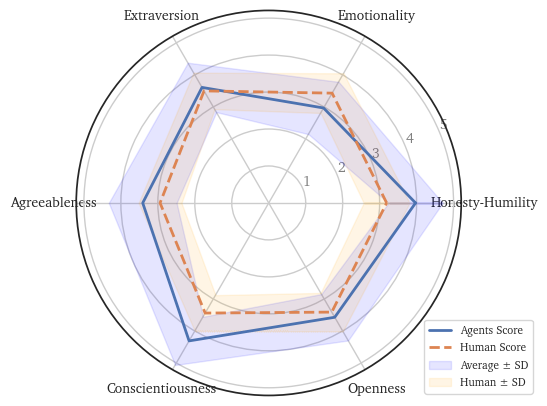}
  \caption{HEXACO-PI-R 100 survey results for PopCensus using Phi-4.}
  \label{fig:pir_phi4}
  \end{minipage}
\end{figure}

To determine whether these consistent personality profiles represent stable characteristics of the agents' personas or are artifacts of the LLM's processing or biases, we repeated the HEXACO-PI-R 100 questionnaire using the same set of personas but with different models, including: Meta's Llama3.2 (3B), Anthropic's Claude 3.7 Sonnet\footnote{2025-02-19}\cite{athropic_claude}, and Microsoft's Phi-4 (14B), the results are presented in Figures \ref{fig:pir_llama}-\ref{fig:pir_phi4}. 


\begin{table}[htbp]
    \centering
    \begin{tabular}{lcccc}
        \toprule
        \textbf{Trait} & \textbf{GPT-4} & \textbf{Sonnet} & \textbf{Phi4} & \textbf{Llama 3.2} \\
        \midrule
        Honesty-Humility   & $-0.814^{*}$ & $-0.768^{*}$ & $-0.851^{*}$ & $-0.214^{*}$ \\
        Emotionality       & $0.686^{*}$  & $0.734^{*}$  & $0.557^{*}$  & $0.117(p=0.039)$ \\
        Extraversion       & $-0.833^{*}$ & $-0.773^{*}$ & $-0.840^{*}$ & $-0.231^{*}$ \\
        Agreeableness      & $-0.873^{*}$ & $-0.868^{*}$ & $-0.882^{*}$ & $-0.360^{*}$ \\
        Conscientiousness  & $-0.934^{*}$ & $-0.928^{*}$ & $-0.896^{*}$ & $-0.368^{*}$ \\
        Openness           & $-0.447^{*}$ & $-0.433^{*}$ & $-0.278^{*}$ & $-0.013(p=0.819)$ \\
        \midrule
        Mean (abs) Average       & $0.765$ & $0.751$ & $0.712$ & $0.217$ \\
        \bottomrule
    \end{tabular}
    \caption{Pearson correlations (r) between lexical analysis and HEXACO-PI-R-100 across different LLMs. Note *$p<.001$.}
    \label{tab:hexaco_llm_correlations}
\end{table}

Table \ref{tab:hexaco_llm_correlations} shows the Pearson correlation coefficients for the four models surveyed.  In this cross-model comparison, we can see that the results from the Sonnet survey, correlates well with the original GPT-4 lexical analysis, with a mean average correlation of 0.752, closely followed by Phi-4's results (0.717).  In fact, it is notable that Phi-4 had a stronger correlation with the GPT-4 lexical study on Honest-Humility, Extraversion and Agreeableness, than GPT-4's results (overall mean correlation 0.765).  Of course, it should be noted that Phi4 was distilled from GPT-4.

% spearman's r for the same results.
\begin{comment}
\begin{table}[htbp]
    \centering
    \label{tab:hexaco_llm_correlations}
    \begin{tabular}{lcccc}
        \toprule
        \textbf{Trait} & \textbf{GPT-4} & \textbf{Sonnet} & \textbf{Phi4} & \textbf{Llama 3.2} \\
        \midrule
        Honesty-Humility   & $-0.661^{*}$ & $-0.683^{*}$ & $-0.721^{*}$ & $-0.201^{*}$ \\
        Emotionality       & $0.678^{*}$  & $0.758^{*}$  & $0.586^{*}$  & $0.127(p=0.025)$ \\
        Extraversion       & $-0.830^{*}$ & $-0.784^{*}$ & $-0.839^{*}$ & $-0.232^{*}$ \\
        Agreeableness      & $-0.900^{*}$ & $-0.889^{*}$ & $-0.860^{*}$ & $-0.349^{*}$ \\
        Conscientiousness  & $-0.901^{*}$ & $-0.882^{*}$ & $-0.865^{*}$ & $-0.368^{*}$ \\
        Openness           & $-0.365^{*}$ & $-0.352^{*}$ & $-0.241^{*}$ & $-0.006(p=0.911)$ \\
        \bottomrule
        \multicolumn{5}{l}{\footnotesize \textit{Note.} *$p<.001$.}
    \end{tabular}
    \caption{Spearman correlations (r) between lexical analysis and HEXACO-PI-R-100 across different LLMs.}
\end{table}
\end{comment}

\subsection{Second Population}

In personality psychology, the validation of lexical studies traditionally involves demonstrating factor replication and consistency.  Indeed, this has been one of the strengths of the HEXACO framework; its six-factor structure emerges across multiple languages, lending credibility to the model.

\begin{figure}[!h]
\centering
  \includegraphics[width=0.8\textwidth]{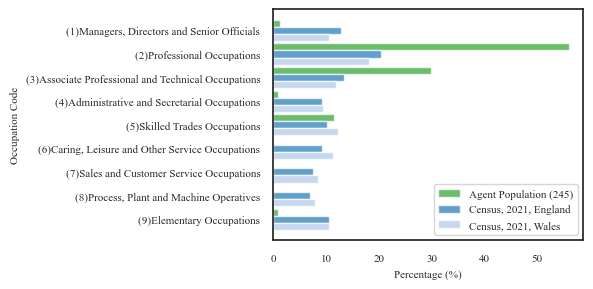}
  \caption{Agent population (PopProfessional) broken down by top-level OSC2020 Classifications against 2021 Census for England and Wales.}
  \label{fig:occupations_pop_a}
\end{figure}

To examine the robustness in our findings, we repeated the experiment with a different, less curated, population (PopProfessional).  This population was created using GPT-4 to generate batches of agents. The prompt used to create the biographies was similar to that used for PopCensus, except it used the term `profession' instead of `occupation'.  This resulted in a skewed distribution of job titles towards professional roles as presented in Figure \ref{fig:occupations_pop_a}.  This population had 83 unique occupations, which included 20 architects, 20 entrepreneurs and 15 graphic designers, and was therefore inherently significantly less diverse, as evidenced by the number of adjectives with a standard deviation of 0 which increased from 7 to 44 for this population.  The terms upon which the agents all returned the same response were: \textit{abusive, bigoted, deceitful, dishonest, underhanded, adulterous, blasphemous, conscienceless, destructive, double-faced, dull-witted, elfish, false-hearted, flammable, fraudulent, hoggish, homicidal, ill-willed, illiterate, inhuman, inhumane, knavish, larcenous, lewd, malevolent, malicious, murderous, overviolent, sadistic, slanderous, thick-witted, treacherous, truthless, unadulterous, unchaste, uncivilized, underwitted, unethical, unfaithful, unprincipled, untrustworthy, untruthful, violent}. Most ($\sim$84\%) of these terms are commonly associated with the Dark Triad~\cite{britannica} personality traits: Machiavellianism, narcissism, and psychopathy.  Traits which would be undesirable in large commercial propriety systems, and as such are likely `trained out' of models prior to deployment, hence the rejection of the notions by all agents.

Across factor solutions ranging from 5 to 12, the eight-factor solution yielded the highest average Cronbach's alpha of 0.66, compared to 0.73 for the previously analysed population (PopCensus).  The top-loaded adjectives for the 8-factor solution are presented in Appendix \ref{app_8_results}.   Semantic similarity between terms within each factor for PopProfessional produced results comparable to those obtained from PopCensus.

Factor 1 (\textbf{Unconscientiousness}) contrasts adjectives along the dimension \textit{meticulous} (-0.55) and \textit{punctual} (-0.55) verses \textit{haphazard} (0.73) and \textit{scatterbrained} (0.66), describing a lack of conscientiousness.  19 of the 30 top-loaded terms were also found in the 30 top-loaded terms in PopCensus' factor 4 (Unconscientiousness).  Factor 2 (\textbf{Introversion}) primarily contains terms reflecting a lack of Extraversion, such as \textit{withdrawn} (0.53), \textit{aloof} (0.51) and \textit{impersonal} (0.49), with no negatively loaded terms in top 30 adjectives.  26 of the 30 top-loaded terms were also found in the top-loadings for PopCensus' factor 3 (Introversion). Factor 3 (\textbf{Unconciliatory}) includes descriptors of interpersonal style, contrasting \textit{brusque} and \textit{unaccomodating} with \textit{forgiving} and \textit{ladylike}. Factor 4 (\textbf{Conventional}) describes a dimension from \textit{traditional} and \textit{unadaptable} to \textit{venturous} and \textit{unconventional}, which aligns somewhat with PopCensus' factor 8 (Unheroic). Factor 5 (\textbf{Extravagant}) includes terms along a dimension of \textit{lavish} and \textit{chatty} to \textit{extravagant} and \textit{unassuming}, and Factor 6 (\textbf{Emotionality}) describes terms related to \textit{sentimental} and \textit{emotional} versus \textit{calculating} and \textit{businesslike}. Factor 7 (\textbf{Heroic}) contrasts \textit{heroic} and \textit{scrappy} versus \textit{artistic} and \textit{fussy}.  This factor does not align specifically to any one HEXACO dimension but is similar to factor 9 (Unartistic) recovered from PopCensus' responses. Factor 8 (\textbf{Unscholarly}) includes terms along a dimension related to Intellect, contrasting \textit{scholarly} and \textit{ineloquent} verses \textit{bookish} and \textit{cerebral} descriptors.  17 of the 30 top-loaded terms were also recovered as part of PopCensus' 6 (Unscholarly).

\begin{figure}[!h]
  \centering
  \begin{minipage}{0.49\textwidth}
  \includegraphics[width=\textwidth]{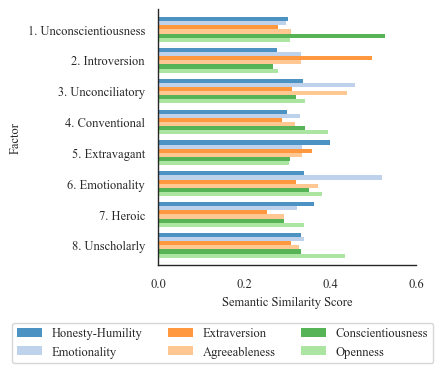}
  \caption{Symmetric Semantic Similarity values for factors in 8-factor solution (PopProfessional).\newline}
  \label{fig:pop_a_similarities}
  \end{minipage}
  \hfill
  \begin{minipage}{0.49\textwidth}
  \includegraphics[width=\textwidth]{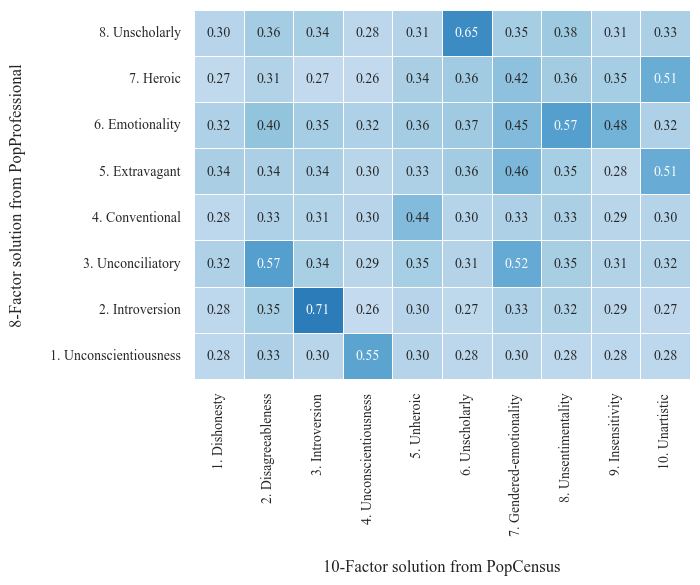}
  \caption{Symmetric Semantic Similarity between factors in PopCensus' 10-factor solution and PopProfessional's 8-factor solution.}
  \label{fig:heatmap}
  \end{minipage}
\end{figure}

The Symmetric Semantic Similarity between factors derived from PopProfessional and the HEXACO dimensions are presented in Figure \ref{fig:pop_a_similarities}. Factors 1, 2, 6 and 8 show notable semantic alignment with the HEXACO dimensions of Contentiousness, Extraversion, Emotionality and Openness. Factor 3 appears to mix terms typically associated with Agreeableness and Emotionality, while Factors 4, 5 and 7 do not have a strong alignment to any specific HEXACO dimension.

These results highlight both similarities and notable differences in recovered structures across populations, offering further insights into the replicability and robustness of the underlying personality dimensions of the agents identified through lexical analysis.

Figure \ref{fig:heatmap} shows the Symmetric Semantic Similarity between PopCensus' 10-factor solution and PopProfessional's 8-factor solution.  These figures show that there is a stronger similarity between the factors recovered for the agent populations, than between either agent population and HEXACO.  Most notably, both population's Introversion and Unscholarly factors align well. 

However, PopCensus' Dishonesty factor did not match well with any of PopProfessional's factors.  During PCA on PopProfessional's results, terms related to honesty-humility were not recovered, which was attributed to an issue in persona generation.  Although instructed to provide three personality facts, two positive and one negative, the LLM tended to provide softer negative facts, for example `bad time management' and `struggles with work-life balance'.  This was rectified for PopCensus, resulting in a broader range of negative personality facts being included in the biographies.  This small adjustment to the generation of the population is the most likely reason why PCA solutions for that population often recovered an Honesty-Humility related factor first, and why the terms in those factors weighted so highly.

In summary, results from the second agent population (PopProfessional) reinforce the replicability and robustness of several core dimensions, particularly Introversion, Unscholarly, Disagreeableness, Unconscientiousness, and Emotionality factors. Nevertheless, notable differences emerged, highlighting how subtle variations in agent biography generation can influence the recovered structure, notably the absence of Honesty-Humility. Consequently, careful attention to the construction and diversity of agent personas is essential when using agents in lexical-based personality research to ensure more representative results.

\subsection{Persona Consistency}

LLMs frequently generate inconsistent or incomplete responses resulting from hallucinations and attention limitations~\cite{ji2023}, potentially affecting research reliability. An agent's ability to maintain a consistent persona throughout a survey is critical for valid outcomes.

To assess agent consistency, responses to 342 antonym adjective pairs (e.g., \textit{kind} vs. \textit{unkind}) were analysed. A consistency score (0 to 1) quantified how consistently each agent endorsed one adjective while rejecting its antonym; scores closer to 1 indicated higher consistency. Scores of 0.875 and 0.75 represented one-level and two-level discrepancies, respectively.  Consistency scores across pairs ranged from 0.06 to 0.99, with 83\% achieving scores of at least 0.75 (indicating strong consistency).

\paragraph{Semantic Generalisation} The lowest-scoring antonym pair was \textit{derogatory}-\textit{underogatory} (0.06). Over 80\% of agents rated both adjectives identically (`Extremely Inaccurate'), suggesting a misunderstanding of \textit{underogatory}, an archaic term\footnotemark{}. This demonstrates semantic opacity where the agents struggled to infer meanings of unfamiliar compound words despite recognisable morphology.
\footnotetext{Oxford English Dictionary's only evidence for \textit{underogatory} is from 1659, in the writing of Robert Boyle, natural philosopher.}

\paragraph{Overly Rigid Gender Bias}\textit{ladylike}-\textit{unladylike} scored second-lowest (0.289), driven by agents dismissing gender-specific descriptors. Nearly all inconsistent responders (199/200) were male (rating `Extremely Inaccurate' for both terms), with mirrored patterns for \textit{manly}-\textit{unmanly} among female agents.  Analysis indicated rigid binary gender encodings were present within the LLM, as less explicitly gendered pairs (e.g., \textit{assertive}-\textit{unassertive}, 0.838) showed significantly higher consistency, demonstrating the presence of strong gender biases likely enforced by outdated (e.g. terms such as \textit{unmaidenly}, \textit{ungentlemanly} and \textit{overwomanly}) or stereotyped vocabulary.

\paragraph{Positive Reframing} Low consistency (0.309) for \textit{impressionable}-\textit{unimpressionable} arose from agents reframing contradictory adjectives positively by adapting the context. Agents inconsistently endorsed both terms, aligning each with favourable attributes, in this case equating \textit{impressionable} as being open to learning and new ideas, and \textit{unimpressionable} as being steadfast and not easily swayed by others. Similar reframing was evident with pairs such as \textit{docile}-\textit{indocile}, supporting previous findings of a tendency towards socially desirable responses in generative agents~\cite{petrov2024}.

\subsubsection{Biography Length vs. Agent Consistency}

\begin{figure}[!ht]
\centering
  \includegraphics[width=0.5\textwidth]{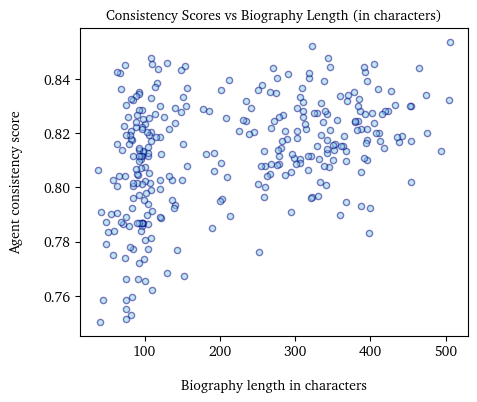}
  \caption{Agent Consistency Scores vs. Biography Length.}
  \label{fig:bio_length}
\end{figure}

Agent consistency was calculated as the mean of individual consistency scores, with an overall average of 0.812 (range: 0.75–0.85). Analysis revealed a relationship between biography length (total character count in hobbies/interests and personality facts) and agent consistency (Figure \ref{fig:bio_length}). Longer, richer biographies generally correlated with higher consistency and lower variance in performance.

The lowest-performing agents (e.g. agent \#113) had notably brief biographies, while the highest-performing agents (e.g. agent \#210) featured more detailed and nuanced character descriptions.  Both agent biographies are presented in Appendix \ref{app_prompts}. This suggests richer biographical context enhances agent consistency, likely by providing more extensive personality anchors for responses.

Overall, while generative agents demonstrated strong persona consistency across most antonym pairs, specific issues such as semantic opacity, rigid gender biases, and positive reframing indicate  limitations in agent response patterns. 

Interestingly, the positive correlation between biography length and consistency scores emphasises that richer, more detailed, less contradictory character biographies significantly enhance agent consistency, providing a broader set of clearer persona artefacts to elicit better responses. This aligns with other findings where adding more contextual information into the prompt elicits better performance: Miotto \textit{et al.}~\cite{miotto2022} found that adding the history of previous responses improved the results.  Huang \textit{et al.}~\cite{huang2024} used an expanded format of the Likert scale~\cite{likert1932}, presenting each response option as a complete sentence, reporting that the expanded responses showed higher convergence with the human participants' data, than the standard Likert ratings.  

Therefore, careful construction and specification of agent biographies is recommended to achieve optimal consistency, improving reliability and validity in studies employing generative agents.  This strong evidence that persona construction is a key aspect for maximising fidelity within agent-based simulations.

\section{Conclusions}

To our knowledge this study represents the first replication of the HEXACO lexical experiment using LLM-powered agents, providing insights into both the capabilities and limitations of using agents as proxies for human participants in personality research.  Our investigation revealed three key findings:

\textbf{Recoverable but distinct personality structure} While a coherent personality structure was recovered from the agent responses, it differs from the human-derived model.  The optimal 10-factor solution demonstrated strong internal consistency (average Cronbach's alpha = 0.73) and meaningful semantic clustering but deviated significantly from the traditional 6-factor HEXACO model.  Most notably, agents fragmented broad human dimensions into more specific facets - Emotionality split across gendered-emotionality, sensitivity, and sentimentality factors, whilst Openness separated into scholarly, heroic and artistic dimensions.  This suggests that LLMs organise personality concepts through different semantic associations than humans, reflecting the statistical patterns in their training data rather than genuine psychological constructs.

\textbf{Model-dependency} Cross-model validation revealed variability in personality profiling across different LLMs. Whilst GPT-4, Sonnet and Phi-4 showed strong convergent validity (mean correlations of 0.765, 0.752 and 0.717 respectively), the smaller Llama 3.2 model demonstrated significantly weaker alignment, highlighting model-specific trends/patterns/biases in personality representation. This variability underscores that apparent personality consistency reflects characteristics of both the specific LLM architecture and the agent population, rather than stable psychological traits.

\textbf{Population curation} The comparison between PopCensus and PopProfessional demonstrates how subtle variations in agent generation profoundly impact recovered personality structures.  The less diverse population failed to recover a Honesty-Humility factor and showed reduced overall reliability (Cronbach's alpha = 9.66 `v' 0.73), showing that representative agent populations require careful, systematic construction to avoid biased or incomplete personality representations.

\subsection{Considerations for Generative AI researchers}

\textbf{}A notable juxtaposition revealed by our study is that generative agents strongly rely on general, broadly-stereotyped associations to assign appropriate personality traits based on the given biography (e.g., software engineers as introverted), yet often fail to  include relevant, contextually specific experiences (e.g., a healthcare worker referencing COVID-19 experiences). This suggests that while current architectures are effective at leveraging stereotypes encoded in their training data, they struggle with integrating nuanced, situationally-specific events (akin to lived experiences) into their responses unless explicitly instructed. This also aligns with previous findings ~\cite{Tan_2025, cheng2023markedpersonasusingnatural, liu2024evaluatinglargelanguagemodel} that suggest that social stereotypes, which appear frequently in the training data, provide a stronger signal compared to nuanced, situational details, which tend to be less discoverable in the data. This results in models being less flexible with incongruent personas, i.e. personas which defy the model's stereotypical expectations, and defaulting to a `typical' demographic responses rather than maintaining conflicting traits. Future research could consider refining prompting strategies, improving biography creation, or enhancing model architectures to enrich stereotypes with situationally specific contexts, thereby achieving more realistic, varied, and contextually aware agent responses.  Additionally, exploring techniques to evaluate how strongly a model's latent knowledge links certain attributes to personas.

\subsection{Implications for social science research}

\textbf{Methodological limitations:} Our results suggest that standard psychometric instruments like HEXACO-PI-R, designed for human introspection and stable identity, are inappropriate for LLM assessment.  The agents' inability to fully replicate human personality structures via lexical analysis, combined with their tendency towards socially desirable responses coupled with rigid adherence to safety training, indicates that direct translation of human psychological methods to generative agents requires theoretical and methodical reconsideration.

\textbf{Design requirements:} Successful agent-based personality research requires sophisticated population curation, with more detailed biographies correlating with response consistency. The emergence of novel personality dimensions unique to agents, such as "Unheroic" and "Unartistic", suggest that agent-based research may require entirely new theoretical frameworks rather than adaptions of existing human models.

\textbf{Validity constraints:} While agents can produce statistically reliable personality-like responses, our research suggests it is not meaningful to ascribe genuine personality traits to LLMs, in line with findings of Bender \textit{et al.}~\cite{bender2021}, Pellert \textit{et al.}~\cite{pellert2024}; LLMs generate learnt linguistic associations presenting statistically plausible but ultimately shallow representations of human personality stereotypes rather than authentic psychological constructs.  This has implications for the usefulness of LLMs in social science simulations, highlighting that when simulating relationships between personal attributes and personality, the LLM draw on stereotypical associations rather than replicating the nuanced relationship observed in humans.

"Such corpora contain sediments of the personalities, values, beliefs, and biases of the countless human authors of these texts, which LLMs learn through a complex training process"~\cite{pellert2024}.

\subsection{Further Work}

\paragraph{Framework development} New theoretical frameworks specifically designed for agent-based personality research are required.  Direct application of human psychological models does not fully capture the unique traits exhibited by agents.  Future frameworks should incorporate shorter, less token intensive versions of personality tests, specifically designed to accommodate limitations of models' vocabulary. Tests should emphasise comparative measures among agents rather than comparisons to human standards.

\paragraph{Model convergence} Further exploration of model convergence for generative agent personalities is required.  Prior work has noted a correlation between personality stability and factors such as model complexity and fine-tuning processes. Our observations suggest similar personality structures emerge in models of comparable size and sophistication, as well as models distilled from larger parent models.  Future studies should investigate whether these observations indicate a convergence in latent stereotyped personas within LLMs.

\paragraph{Behavioural alignment} Reliable personality assessment of agents should ideally allow us to predict their behaviours and decisions.  Future research is required to determine if stable agent populations indeed behave consistently according to their personality traits, particularly exploring how closely these traits predict behaviours similar to humans with comparable personality profiles, acknowledging constraints imposed by structural differences.  Additionally, longitudinal studies assessing agent personalities over extended simulation could provide insights into personality stability within dynamic environments.

\subsection*{Data Availability}
The full dataset (populations, responses, aggregated data, and code) is publicly available at \newline https://github.com/alan-turing-institute/hexaco-rep-public

\subsection*{Acknowledgements}
We are grateful to Prof. Tim Watson, Ms Sandra Clarke, and Prof. Paul Martin for their valuable comments on earlier drafts of this paper.

\newpage
\bibliographystyle{IEEEtran} % to be used for final paper!
\bibliography{bibliography}

\begin{thebibliography}{10}
\providecommand{\url}[1]{#1}
\csname url@samestyle\endcsname
\providecommand{\newblock}{\relax}
\providecommand{\bibinfo}[2]{#2}
\providecommand{\BIBentrySTDinterwordspacing}{\spaceskip=0pt\relax}
\providecommand{\BIBentryALTinterwordstretchfactor}{4}
\providecommand{\BIBentryALTinterwordspacing}{\spaceskip=\fontdimen2\font plus
\BIBentryALTinterwordstretchfactor\fontdimen3\font minus \fontdimen4\font\relax}
\providecommand{\BIBforeignlanguage}[2]{{%
\expandafter\ifx\csname l@#1\endcsname\relax
\typeout{** WARNING: IEEEtran.bst: No hyphenation pattern has been}%
\typeout{** loaded for the language `#1'. Using the pattern for}%
\typeout{** the default language instead.}%
\else
\language=\csname l@#1\endcsname
\fi
#2}}
\providecommand{\BIBdecl}{\relax}
\BIBdecl

\bibitem{gurcan2024}
\BIBentryALTinterwordspacing
O.~Gurcan, ``Llm-augmented agent-based modelling for social simulations: Challenges and opportunities,'' 2024. [Online]. Available: \url{https://arxiv.org/abs/2405.06700}
\BIBentrySTDinterwordspacing

\bibitem{hua2024}
\BIBentryALTinterwordspacing
W.~Hua, L.~Fan, L.~Li, K.~Mei, J.~Ji, Y.~Ge, L.~Hemphill, and Y.~Zhang, ``War and peace (waragent): Large language model-based multi-agent simulation of world wars,'' 2024. [Online]. Available: \url{https://arxiv.org/abs/2311.17227}
\BIBentrySTDinterwordspacing

\bibitem{park2023simulacra}
\BIBentryALTinterwordspacing
J.~S. Park, J.~C. O'Brien, C.~J. Cai, M.~R. Morris, P.~Liang, and M.~S. Bernstein, ``Generative agents: Interactive simulacra of human behavior,'' 2023. [Online]. Available: \url{https://arxiv.org/abs/2304.03442}
\BIBentrySTDinterwordspacing

\bibitem{horton2023}
\BIBentryALTinterwordspacing
J.~J. Horton, ``Large language models as simulated economic agents: What can we learn from homo silicus?'' 2023. [Online]. Available: \url{https://arxiv.org/abs/2301.07543}
\BIBentrySTDinterwordspacing

\bibitem{guo2024survey}
\BIBentryALTinterwordspacing
T.~Guo, X.~Chen, Y.~Wang, R.~Chang, S.~Pei, N.~V. Chawla, O.~Wiest, and X.~Zhang, ``Large language model based multi-agents: A survey of progress and challenges,'' 2024. [Online]. Available: \url{https://arxiv.org/abs/2402.01680}
\BIBentrySTDinterwordspacing

\bibitem{aws_blog}
\BIBentryALTinterwordspacing
I.~Gleiser, ``Llms: the new frontier in generative agent-based simulation,'' October 2024. [Online]. Available: \url{https://aws.amazon.com/blogs/hpc/llms-the-new-frontier-in-generative-agent-based-simulation/}
\BIBentrySTDinterwordspacing

\bibitem{ziems2024}
\BIBentryALTinterwordspacing
C.~Ziems, W.~Held, O.~Shaikh, J.~Chen, Z.~Zhang, and D.~Yang, ``Can large language models transform computational social science?'' \emph{Computational Linguistics}, vol.~50, no.~1, pp. 237--291, 03 2024. [Online]. Available: \url{https://doi.org/10.1162/coli\_a\_00502}
\BIBentrySTDinterwordspacing

\bibitem{lazer2020}
\BIBentryALTinterwordspacing
D.~M.~J. Lazer, A.~Pentland, D.~J. Watts, S.~Aral, S.~Athey, N.~Contractor, D.~Freelon, S.~Gonzalez-Bailon, G.~King, H.~Margetts, A.~Nelson, M.~J. Salganik, M.~Strohmaier, A.~Vespignani, and C.~Wagner, ``Computational social science: Obstacles and opportunities,'' \emph{Science}, vol. 369, no. 6507, pp. 1060--1062, 2020. [Online]. Available: \url{https://www.science.org/doi/abs/10.1126/science.aaz8170}
\BIBentrySTDinterwordspacing

\bibitem{hagendorff2024}
\BIBentryALTinterwordspacing
T.~Hagendorff, I.~Dasgupta, M.~Binz, S.~C.~Y. Chan, A.~Lampinen, J.~X. Wang, Z.~Akata, and E.~Schulz, ``Machine psychology,'' 2024. [Online]. Available: \url{https://arxiv.org/abs/2303.13988}
\BIBentrySTDinterwordspacing

\bibitem{li2024quantifyingaipsychology}
\BIBentryALTinterwordspacing
Y.~Li, Y.~Huang, H.~Wang, X.~Zhang, J.~Zou, and L.~Sun, ``Quantifying ai psychology: A psychometrics benchmark for large language models,'' 2024. [Online]. Available: \url{https://arxiv.org/abs/2406.17675}
\BIBentrySTDinterwordspacing

\bibitem{pellert2024}
\BIBentryALTinterwordspacing
M.~Pellert, C.~M. Lechner, C.~Wagner, B.~Rammstedt, and M.~Strohmaier, ``Ai psychometrics: Assessing the psychological profiles of large language models through psychometric inventories,'' \emph{Perspectives on Psychological Science}, vol.~19, no.~5, pp. 808--826, 2024, pMID: 38165766. [Online]. Available: \url{https://doi.org/10.1177/17456916231214460}
\BIBentrySTDinterwordspacing

\bibitem{petrov2024}
\BIBentryALTinterwordspacing
N.~B. Petrov, G.~Serapio-García, and J.~Rentfrow, ``Limited ability of llms to simulate human psychological behaviours: a psychometric analysis,'' 2024. [Online]. Available: \url{https://arxiv.org/abs/2405.07248}
\BIBentrySTDinterwordspacing

\bibitem{bail2024}
\BIBentryALTinterwordspacing
{C.A. Bail}, ``Can generative ai improve social science?'' \emph{Proc. Natl. Acad. Sci. U.S.A. 121 (21) e2314021121}, 2024. [Online]. Available: \url{https://doi.org/10.1073/pnas.2314021121}
\BIBentrySTDinterwordspacing

\bibitem{argyle2023}
L.~P. Argyle, E.~C. Busby, N.~Fulda, J.~R. Gubler, C.~Rytting, and D.~Wingate, ``Out of one, many: Using language models to simulate human samples,'' \emph{Political Analysis}, vol.~31, no.~3, p. 337–351, 2023.

\bibitem{simmons2023}
\BIBentryALTinterwordspacing
G.~Simmons and C.~Hare, ``Large language models as subpopulation representative models: A review,'' 2023. [Online]. Available: \url{https://arxiv.org/abs/2310.17888}
\BIBentrySTDinterwordspacing

\bibitem{bender2021}
\BIBentryALTinterwordspacing
E.~M. Bender, T.~Gebru, A.~McMillan-Major, and S.~Shmitchell, ``On the dangers of stochastic parrots: Can language models be too big?'' in \emph{Proceedings of the 2021 ACM Conference on Fairness, Accountability, and Transparency}, ser. FAccT '21.\hskip 1em plus 0.5em minus 0.4em\relax New York, NY, USA: Association for Computing Machinery, 2021, p. 610–623. [Online]. Available: \url{https://doi.org/10.1145/3442188.3445922}
\BIBentrySTDinterwordspacing

\bibitem{marcus2020}
\BIBentryALTinterwordspacing
G.~Marcus, ``The next decade in ai: Four steps towards robust artificial intelligence,'' 2020. [Online]. Available: \url{https://arxiv.org/abs/2002.06177}
\BIBentrySTDinterwordspacing

\bibitem{shane2005}
\BIBentryALTinterwordspacing
S.~Frederick, ``Cognitive reflection and decision making,'' \emph{Journal of Economic Perspectives}, vol.~19, no.~4, p. 25–42, December 2005. [Online]. Available: \url{https://www.aeaweb.org/articles?id=10.1257/089533005775196732}
\BIBentrySTDinterwordspacing

\bibitem{ericson1981}
\BIBentryALTinterwordspacing
T.~D. Erickson and M.~E. Mattson, ``From words to meaning: A semantic illusion,'' \emph{Journal of Verbal Learning and Verbal Behavior}, vol.~20, no.~5, pp. 540--551, 1981. [Online]. Available: \url{https://www.sciencedirect.com/science/article/pii/S0022537181901651}
\BIBentrySTDinterwordspacing

\bibitem{hagendorff2023}
\BIBentryALTinterwordspacing
T.~Hagendorff, ``Machine psychology: Investigating emergent capabilities and behavior in large language models using psychological methods,'' 2023. [Online]. Available: \url{https://arxiv.org/abs/2303.13988v1}
\BIBentrySTDinterwordspacing

\bibitem{macmillan2924}
\BIBentryALTinterwordspacing
{O. Macmillan-Scott, M. Musolesi}, ``(ir)rationality and cognitive biases in large language models.'' 2024. [Online]. Available: \url{https://doi.org/10.1098/rsos.240255}
\BIBentrySTDinterwordspacing

\bibitem{kahneman1972}
\BIBentryALTinterwordspacing
D.~Kahneman and A.~Tversky, ``Subjective probability: A judgment of representativeness,'' \emph{Cognitive Psychology}, vol.~3, no.~3, pp. 430--454, 1972. [Online]. Available: \url{https://www.sciencedirect.com/science/article/pii/0010028572900163}
\BIBentrySTDinterwordspacing

\bibitem{miotto2022}
\BIBentryALTinterwordspacing
M.~Miotto, N.~Rossberg, and B.~Kleinberg, ``Who is gpt-3? an exploration of personality, values and demographics,'' 2022. [Online]. Available: \url{https://arxiv.org/abs/2209.14338}
\BIBentrySTDinterwordspacing

\bibitem{ashton2009}
{M. C. Ashton, and K. Lee}, ``The hexaco-60: a short measure of the major dimensions of personality.'' \emph{Journal of personality assessment, 91(4), 340–345.}, 2009.

\bibitem{safdari2023}
\BIBentryALTinterwordspacing
M.~Safdari, G.~Serapio-García, C.~Crepy, S.~Fitz, P.~Romero, L.~Sun, M.~Abdulhai, A.~Faust, and M.~Matarić, ``{Personality Traits in Large Language Models (v1)},'' 2023. [Online]. Available: \url{https://arxiv.org/pdf/2307.00184v1}
\BIBentrySTDinterwordspacing

\bibitem{goldberg1999}
L.~Goldberg, ``{A Broad-Bandwidth, Public Domain, Personality Inventory Measuring the Lower-Level Facets of Several Five-Factor Models},'' p. 7–28, 1999.

\bibitem{john1999}
O.~P. John and S.~Srivastava, ``{The Big Five Trait taxonomy: History, measurement, and theoretical perspectives.}'' p. 102–138, 1999.

\bibitem{gupta2024}
\BIBentryALTinterwordspacing
A.~Gupta, X.~Song, and G.~Anumanchipalli, ``Self-assessment tests are unreliable measures of llm personality,'' 2024. [Online]. Available: \url{https://arxiv.org/abs/2309.08163}
\BIBentrySTDinterwordspacing

\bibitem{johnson2014}
\BIBentryALTinterwordspacing
J.~A. Johnson, ``{Measuring thirty facets of the Five Factor Model with a 120-item public domain inventory: Development of the IPIP-NEO-120},'' \emph{Journal of Research in Personality}, vol.~51, pp. 78--89, 2014. [Online]. Available: \url{https://www.sciencedirect.com/science/article/pii/S0092656614000506}
\BIBentrySTDinterwordspacing

\bibitem{huang2024-reliability}
\BIBentryALTinterwordspacing
J.-t. Huang, W.~Jiao, M.~H. Lam, E.~J. Li, W.~Wang, and M.~Lyu, ``On the reliability of psychological scales on large language models,'' in \emph{Proceedings of the 2024 Conference on Empirical Methods in Natural Language Processing}, Y.~Al-Onaizan, M.~Bansal, and Y.-N. Chen, Eds.\hskip 1em plus 0.5em minus 0.4em\relax Miami, Florida, USA: Association for Computational Linguistics, Nov. 2024, pp. 6152--6173. [Online]. Available: \url{https://aclanthology.org/2024.emnlp-main.354/}
\BIBentrySTDinterwordspacing

\bibitem{huang2024}
\BIBentryALTinterwordspacing
M.~Huang, X.~Zhang, C.~Soto, and J.~Evans, ``Designing llm-agents with personalities: A psychometric approach,'' 2024. [Online]. Available: \url{https://arxiv.org/abs/2410.19238}
\BIBentrySTDinterwordspacing

\bibitem{dewinter2024}
\BIBentryALTinterwordspacing
J.~C. {de Winter}, T.~Driessen, and D.~Dodou, ``The use of chatgpt for personality research: Administering questionnaires using generated personas,'' \emph{Personality and Individual Differences}, vol. 228, p. 112729, 2024. [Online]. Available: \url{https://www.sciencedirect.com/science/article/pii/S0191886924001892}
\BIBentrySTDinterwordspacing

\bibitem{rammstedt2007}
\BIBentryALTinterwordspacing
B.~Rammstedt and O.~P. John, ``Measuring personality in one minute or less: A 10-item short version of the big five inventory in english and german,'' \emph{Journal of Research in Personality}, vol.~41, no.~1, pp. 203--212, 2007. [Online]. Available: \url{https://www.sciencedirect.com/science/article/pii/S0092656606000195}
\BIBentrySTDinterwordspacing

\bibitem{ashton2007}
\BIBentryALTinterwordspacing
M.~C. Ashton and K.~Lee, ``Empirical, theoretical, and practical advantages of the hexaco model of personality structure,'' \emph{Personality and Social Psychology Review}, vol.~11, no.~2, pp. 150--166, 2007, pMID: 18453460. [Online]. Available: \url{https://doi.org/10.1177/1088868306294907}
\BIBentrySTDinterwordspacing

\bibitem{ashton2004}
\BIBentryALTinterwordspacing
{M. C. Ashton, K. Lee, and L. R. Goldberg}, ``A hierarchical analysis of 1,710 english personality-descriptive adjectives.'' \emph{Journal of Personality and Social Psychology, 87(5), 707–721.}, 2004. [Online]. Available: \url{https://doi.org/10.1037/0022-3514.87.5.707}
\BIBentrySTDinterwordspacing

\bibitem{hilliard2024}
\BIBentryALTinterwordspacing
A.~Hilliard, C.~Munoz, Z.~Wu, and A.~S. Koshiyama, ``Eliciting personality traits in large language models,'' 2024. [Online]. Available: \url{https://arxiv.org/abs/2402.08341}
\BIBentrySTDinterwordspacing

\bibitem{jiang2024}
\BIBentryALTinterwordspacing
H.~Jiang, X.~Zhang, X.~Cao, C.~Breazeal, D.~Roy, and J.~Kabbara, ``Personallm: Investigating the ability of large language models to express personality traits,'' 2024. [Online]. Available: \url{https://arxiv.org/abs/2305.02547}
\BIBentrySTDinterwordspacing

\bibitem{mercer}
\BIBentryALTinterwordspacing
S.~Mercer, ``Welcome to willowbrook - the simulated society built by generative agents,'' 2023. [Online]. Available: \url{https://cetas.turing.ac.uk/publications/welcome-willowbrook}
\BIBentrySTDinterwordspacing

\bibitem{serapiogarcía2024}
\BIBentryALTinterwordspacing
G.~Serapio-García, M.~Safdari, C.~Crepy, L.~Sun, S.~Fitz, P.~Romero, M.~Abdulhai, A.~Faust, and M.~Matarić, ``Personality traits in large language models,'' 2024. [Online]. Available: \url{https://arxiv.org/abs/2307.00184}
\BIBentrySTDinterwordspacing

\bibitem{zhang2018}
\BIBentryALTinterwordspacing
S.~Zhang, E.~Dinan, J.~Urbanek, A.~Szlam, D.~Kiela, and J.~Weston, ``Personalizing dialogue agents: {I} have a dog, do you have pets too?'' in \emph{Proceedings of the 56th Annual Meeting of the Association for Computational Linguistics (Volume 1: Long Papers)}, I.~Gurevych and Y.~Miyao, Eds.\hskip 1em plus 0.5em minus 0.4em\relax Melbourne, Australia: Association for Computational Linguistics, Jul. 2018, pp. 2204--2213. [Online]. Available: \url{https://aclanthology.org/P18-1205/}
\BIBentrySTDinterwordspacing

\bibitem{bfi2}
{C. J. Soto, and O. P. John}, ``{The next Big Five Inventory (BFI-2): Developing and assessing a hierarchical model with 15 facets to enhance bandwidth, fidelity, and predictive power.}'' \emph{{Journal of Personality and Social Psychology}}, pp. {117--143}, 2017.

\bibitem{ONS_2022}
\BIBentryALTinterwordspacing
{Office for National Statistics (ONS)}, ``Industry and occupation, england and wales: Census 2021,'' 2022. [Online]. Available: \url{https://www.ons.gov.uk/employmentandlabourmarket/peopleinwork/employmentandemployeetypes/bulletins/industryandoccupationenglandandwales/census2021}
\BIBentrySTDinterwordspacing

\bibitem{ONS_SOC}
\BIBentryALTinterwordspacing
{Office for National Statistics}, ``{Extended Standard Occupational Classification (SOC) 2020},'' 2020. [Online]. Available: \url{https://www.ons.gov.uk/methodology/classificationsandstandards/standardoccupationalclassificationsoc/soc2020}
\BIBentrySTDinterwordspacing

\bibitem{white2023}
\BIBentryALTinterwordspacing
J.~White, Q.~Fu, S.~Hays, M.~Sandborn, C.~Olea, H.~Gilbert, A.~Elnashar, J.~Spencer-Smith, and D.~C. Schmidt, ``A prompt pattern catalog to enhance prompt engineering with chatgpt,'' 2023. [Online]. Available: \url{https://arxiv.org/abs/2302.11382}
\BIBentrySTDinterwordspacing

\bibitem{jung2022}
\BIBentryALTinterwordspacing
J.~Jung, L.~Qin, S.~Welleck, F.~Brahman, C.~Bhagavatula, R.~L. Bras, and Y.~Choi, ``Maieutic prompting: Logically consistent reasoning with recursive explanations,'' 2022. [Online]. Available: \url{https://arxiv.org/abs/2205.11822}
\BIBentrySTDinterwordspacing

\bibitem{condon2021}
\BIBentryALTinterwordspacing
D.~Condon, J.~Coughlin, and S.~Weston, ``{Trait Descriptive Adjectives},'' 2021. [Online]. Available: \url{https://doi.org/10.7910/DVN/5T80PF}
\BIBentrySTDinterwordspacing

\bibitem{gpt4azure}
\BIBentryALTinterwordspacing
E.~Boyd, ``Introducing gpt-4 in azure openai service,'' 2023. [Online]. Available: \url{https://azure.microsoft.com/en-us/blog/introducing-gpt4-in-azure-openai-service/}
\BIBentrySTDinterwordspacing

\bibitem{hendrickson}
\BIBentryALTinterwordspacing
A.~E. Hendrickson and P.~O. White, ``Promax: A quick method for rotation to oblique simple structure,'' \emph{British Journal of Statistical Psychology}, vol.~17, no.~1, pp. 65--70, 1964. [Online]. Available: \url{https://bpspsychub.onlinelibrary.wiley.com/doi/abs/10.1111/j.2044-8317.1964.tb00244.x}
\BIBentrySTDinterwordspacing

\bibitem{saucier1996}
\BIBentryALTinterwordspacing
{Saucier, G., and Goldberg, L. R.}, ``Evidence for the big five in analyses of familiar english personality adjectives.'' \emph{European Journal of Personality, 10(1), 61-77.}, 1996. [Online]. Available: \url{https://doi.org/10.1002/(SICI)1099-0984(199603)10:1<61::AID-PER246>3.0.CO;2-D}
\BIBentrySTDinterwordspacing

\bibitem{jim}
\BIBentryALTinterwordspacing
J.~Frost, ``Cronbach’s alpha: Definition, calculations and example,'' accessed March 2025. [Online]. Available: \url{https://statisticsbyjim.com/basics/cronbachs-alpha/}
\BIBentrySTDinterwordspacing

\bibitem{athropic_claude}
\BIBentryALTinterwordspacing
A.~Team, ``{Claude 3.7 Sonnet and Claude Code},'' 2025. [Online]. Available: \url{https://www.anthropic.com/news/claude-3-7-sonnet}
\BIBentrySTDinterwordspacing

\bibitem{britannica}
\BIBentryALTinterwordspacing
K.~Akre, ``Dark triad,'' accessed 2025. [Online]. Available: \url{https://www.britannica.com/science/dark-triad}
\BIBentrySTDinterwordspacing

\bibitem{ji2023}
\BIBentryALTinterwordspacing
Z.~Ji, N.~Lee, R.~Frieske, T.~Yu, D.~Su, Y.~Xu, E.~Ishii, Y.~J. Bang, A.~Madotto, and P.~Fung, ``{Survey of Hallucination in Natural Language Generation},'' \emph{ACM Comput. Surv.}, vol.~55, no.~12, Mar. 2023. [Online]. Available: \url{https://doi.org/10.1145/3571730}
\BIBentrySTDinterwordspacing

\bibitem{likert1932}
R.~Likert, ``{A technique for the measurement of attitudes},'' \emph{{Archives of Psychology}}, vol. {22 140, 55}, 1932.

\bibitem{Tan_2025}
\BIBentryALTinterwordspacing
B.~C.~Z. Tan and R.~K.-W. Lee, ``Unmasking implicit bias: Evaluating persona-prompted llm responses in power-disparate social scenarios,'' in \emph{Proceedings of the 2025 Conference of the Nations of the Americas Chapter of the Association for Computational Linguistics: Human Language Technologies (Volume 1: Long Papers)}.\hskip 1em plus 0.5em minus 0.4em\relax Association for Computational Linguistics, 2025, p. 1075–1108. [Online]. Available: \url{http://dx.doi.org/10.18653/v1/2025.naacl-long.50}
\BIBentrySTDinterwordspacing

\bibitem{cheng2023markedpersonasusingnatural}
\BIBentryALTinterwordspacing
M.~Cheng, E.~Durmus, and D.~Jurafsky, ``Marked personas: Using natural language prompts to measure stereotypes in language models,'' 2023. [Online]. Available: \url{https://arxiv.org/abs/2305.18189}
\BIBentrySTDinterwordspacing

\bibitem{liu2024evaluatinglargelanguagemodel}
\BIBentryALTinterwordspacing
A.~Liu, M.~Diab, and D.~Fried, ``Evaluating large language model biases in persona-steered generation,'' 2024. [Online]. Available: \url{https://arxiv.org/abs/2405.20253}
\BIBentrySTDinterwordspacing

\end{thebibliography}

\newpage

 % main document

\appendix
\section{Appendix: Prompts}
\label{app_prompts}
The following system-prompt and user-prompt were used to generate the character biographies:
\paragraph{\ttfamily
“You are a character generator AI. \newline
Your responses should be json objects. \newline 
Do not use names of famous people. \newline
Ages should be between 16 and 60. \newline 
Occupations can also include unpaid activities, e.g. student, stay at home mum, job seeker etc.
  \newline
  \newline
“Complete this character bio, where an occupation has already been given:\newline
Full Name: [Full Name]\newline
Age: [Age]\newline
Occupation: {occupation}\newline
Hobbies/Interests: [Hobbies/Interests]\newline
Personality Facts:\newline
- [Positive Fact 1]\newline
- [Positive Fact 2]\newline
- [Negative Fact]”}

\paragraph{}\color{white}-\color{black} % fix this hack!

\noindent Here are some of examples of the biographies created:

\paragraph{\ttfamily
Agent ID \#113:\newline\newline
   "Full Name": "Lee Scott",\newline
   "Age": 18,\newline
   "Occupation": "Personal Trainer",\newline
   "Hobbies/interests": "fitness, nutrition, outdoor sports",\newline
   "Personality Facts": \{\newline
   "Positive Fact 1": "Motivated",\newline
   "Positive Fact 2": "Disciplined",\newline
   "Negative Fact": "Can be overly competitive" \} \newline
}

\paragraph{\ttfamily
Agent ID \#210:\newline\newline
   "Full Name": "Margaret `Maggie' Dupont",\newline
   "Age": 34,\newline
   "Occupation": "Pet carer",\newline
   "Hobbies/interests": "nature photography, bird watching, artisanal soap making, gardening",\newline
   "Personality Facts": \{ \newline
   "Positive Fact 1": "Maggie is extremely patient, a trait that makes her excellent at her job dealing with pets of all temperament.",\newline
   "Positive Fact 2": "She is also known for her kindness and empathy, not just towards animals, but also people, always ready to lend a listening ear.",\newline
   "Negative Fact": "However, Maggie has a manipulative side. She uses her knowledge and understanding of animal behavior to sometimes trick people into doing what she wants, especially if it benefits the animals under her care." \} \newline
}

\paragraph{}\color{white}-\color{black} % fix this hack!

\paragraph{}\color{white}-\color{black} % fix this hack!

\noindent This system prompt and prompt were used to conduct the lexical analysis survey:

\paragraph{\ttfamily
    "You are a character in a simulation.\newline
    Answer the next question in character.\newline
    Here is your character bio (in JSON): \{biography\}\newline
    Please ensure your answer starts with the rating from the scale provided."\newline
    \newline
    "Please indicate using the follow scale how accurately this adjective `\{attribute\}' describes you.
    \newline
    `Extremely Inaccurate', `Very Inaccurate', `Moderately Inaccurate', `Slightly Inaccurate', `Neither Accurate Nor Inaccurate', `Slightly Accurate', `Moderately Accurate', 'Very Accurate', `Extremely Accurate'"
    \newline
}

\paragraph{}\color{white}-\color{black} % fix this hack!

\noindent This system prompt and prompt were used to conduct the HEXACO-PI-R 100 survey:

\paragraph{\ttfamily
    "You are a character in a simulation.\newline
    Answer the next question in character.\newline
    Here is your character bio (in JSON): \{biography\}"\newline
    \newline
    How much do you agree or disagree with the the following statement: `\{pi\_question\}'. 
    Please respond using the following scale: Strongly agree, Agree, Neutral (neither agree nor disagree), Disagree, Strongly disagree.
    Ensure your answer starts with the rating from the scale provided, followed by a short explanation.
}
\paragraph{}\color{white}-\color{black} % fix this hack!

\noindent Where \textit{pi\_question} was taken from the 100-Item Version of the Self-Report Form: \url{https://hexaco.org/hexaco-inventory}

\newpage
\section{Appendix: Factor-5 results}
\label{app_5_results}

\begin{table}[!h]
    \centering
    \renewcommand{\arraystretch}{1}
    \setlength{\tabcolsep}{3pt}
    \begin{tabular}{lrlrlrlrlr}
    \hline
    \multicolumn{10}{c}{Factor} \\
    \multicolumn{2}{c}{(1) Introversion} 
    & \multicolumn{2}{c}{(2) Assertiveness} 
    & \multicolumn{2}{c}{(3) Dishonesty} 
    & \multicolumn{2}{c}{(4) Unconventionality}
    & \multicolumn{2}{c}{(5) Provincial} \\
    \multicolumn{2}{c}{(4.84\% of variance)} 
    & \multicolumn{2}{c}{(4.43\% of variance)} 
    & \multicolumn{2}{c}{(4.16\% of variance)} 
    & \multicolumn{2}{c}{(4.11\% of variance)} 
    & \multicolumn{2}{c}{(2.00\% of variance)} \\
    Adjective & Load & Adjective & Load & Adjective & Load & Adjective & Load & Adjective & Load \\
    \hline
    ungregarious	&	0.79	&	unbold	&	-0.77	&	deceptive	&	0.88	&	strait-laced	&	-0.68	&	homespun	&	0.59	\\
    distant	&	0.78	&	placid	&	-0.75	&	manipulative	&	0.88	&	overrigid	&	-0.62	&	folksy	&	0.57	\\
    uncompanionable	&	0.77	&	mild	&	-0.75	&	sly	&	0.87	&	overscrupulous	&	-0.61	&	unliterary	&	0.57	\\
    aloof	&	0.76	&	soft-spoken	&	-0.74	&	sneaky	&	0.86	&	conventional	&	-0.60	&	unscholarly	&	0.57	\\
    standoffish	&	0.76	&	unobtrusive	&	-0.70	&	devious	&	0.86	&	free-living	&	0.59	&	unstudious	&	0.56	\\
    detached	&	0.74	&	forceful	&	0.69	&	deceitful	&	0.84	&	unconventional	&	0.59	&	sophisticated	&	-0.54	\\
    unpersonable	&	0.73	&	outspoken	&	0.66	&	double-faced	&	0.84	&	unpredictable	&	0.58	&	unbookish	&	0.52	\\
    untalkative	&	0.72	&	overpatient	&	-0.65	&	underhanded	&	0.84	&	spontaneous	&	0.58	&	ultraintellectual	&	-0.50	\\
    unaccessible	&	0.71	&	fierce	&	0.65	&	double-tongued	&	0.83	&	unconstrained	&	0.58	&	old-fashioned	&	0.50	\\
    unapproachable	&	0.71	&	bullish	&	0.63	&	unscrupulous	&	0.82	&	hit-or-miss	&	0.57	&	visionary	&	-0.50	\\
    unsociable	&	0.71	&	unassuming	&	-0.63	&	undeceptive	&	-0.82	&	impulsive	&	0.57	&	rugged	&	0.49	\\
    unsolicitous	&	0.71	&	sharp-tongued	&	0.62	&	scheming	&	0.82	&	overconscientious	&	-0.57	&	overstudious	&	-0.49	\\
    taciturn	&	0.70	&	forbearing	&	-0.62	&	undevious	&	-0.81	&	overneat	&	-0.57	&	unprogressive	&	0.48	\\
    warmthless	&	0.70	&	self-assertive	&	0.62	&	pretenseful	&	0.81	&	unreined	&	0.56	&	cosmopolitan	&	-0.48	\\
    uncommunicative	&	0.69	&	forcible	&	0.61	&	incorrupt	&	-0.81	&	unspontaneous	&	-0.56	&	unprovincial	&	-0.47	\\
    asocial	&	0.69	&	fiery	&	0.61	&	uncandid	&	0.80	&	devil-may-care	&	0.55	&	cerebral	&	-0.45	\\
    seclusive	&	0.69	&	overfierce	&	0.60	&	untrustful	&	0.79	&	footloose	&	0.55	&	abstract	&	-0.45	\\
    undemonstrative	&	0.69	&	retiring	&	-0.59	&	untrustworthy	&	0.78	&	conservative	&	-0.55	&	ultrarefined	&	-0.45	\\
    unsocial	&	0.69	&	competitory	&	0.59	&	self-seeking	&	0.77	&	unconstrainable	&	0.54	&	progressive	&	-0.44	\\
    sociable	&	-0.68	&	hard-nosed	&	0.58	&	evasive	&	0.77	&	distractible	&	0.54	&	unskeptical	&	0.44	\\
    dissocial	&	0.68	&	unaggressive	&	-0.58	&	exploitative	&	0.76	&	wild	&	0.54	&	changeless	&	0.44	\\
    social	&	-0.67	&	forward	&	0.58	&	knavish	&	0.76	&	nonrigid	&	0.54	&	unchanging	&	0.43	\\
    terse	&	0.66	&	gentle-minded	&	-0.57	&	dishonest	&	0.75	&	stringent	&	-0.53	&	conventional	&	0.43	\\
    incongenial	&	0.65	&	leisurely	&	-0.57	&	plain-dealing	&	-0.75	&	rigid	&	-0.53	&	unsophisticated	&	0.43	\\
    impersonal	&	0.65	&	sedate	&	-0.57	&	roguish	&	0.75	&	overcareful	&	-0.53	&	forward-looking	&	-0.42	\\
    hermitish	&	0.65	&	hectic	&	0.56	&	untransparent	&	0.75	&	messy	&	0.53	&	unphilosophical	&	0.41	\\
    withdrawn	&	0.64	&	dominant	&	0.56	&	cagey	&	0.74	&	changeable	&	0.53	&	provincial	&	0.41	\\
    congenial	&	-0.63	&	unargumentative	&	-0.56	&	incorruptible	&	-0.74	&	overstrict	&	-0.53	&	urbane	&	-0.41	\\
    reclusive	&	0.63	&	bossy	&	0.55	&	ingratiatory	&	0.74	&	uninhibited	&	0.53	&	brainy	&	-0.41	\\
    accessible	&	-0.63	&	demanding	&	0.55	&	ingratiating	&	0.74	&	inconfinable	&	0.52	&	scholarly	&	-0.40	\\
    \hline
    \end{tabular}
    \caption{Highest loaded terms recovered from the 5-Factor solution.}
    \label{tab:factor_5}
\end{table}

\newpage{}
\section{Appendix: Weighted Jaccard Similarities}
\label{app_jaccard}

\begin{figure}[!h]
  \centering
  \begin{minipage}[b]{0.45\textwidth}
    \includegraphics[width=\textwidth]{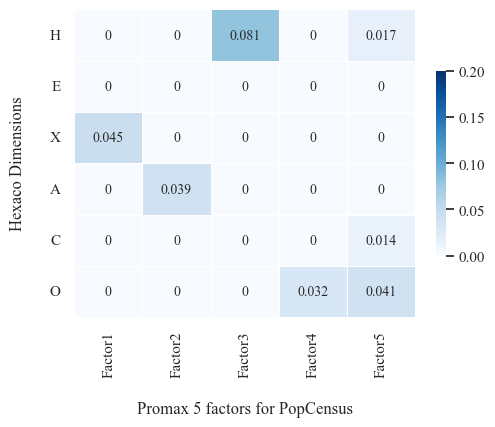}
    \caption{Weighted Jaccard similarity for Promax (5-Factor) results.}
    \label{fig:jaccard_5}
  \end{minipage}
  \hfill
  \begin{minipage}[b]{0.45\textwidth}
    \includegraphics[width=\textwidth]{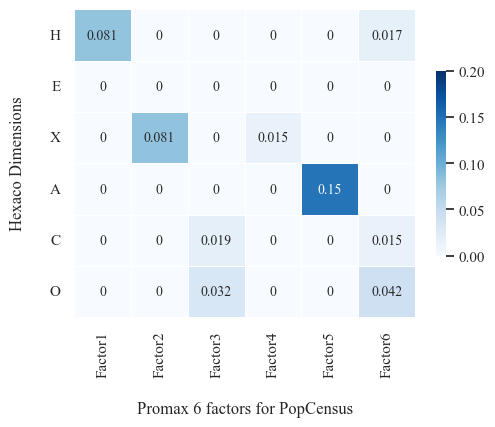}
    \caption{Weighted Jaccard similarity for Promax (6-Factor) results.}
    \label{fig:jaccard_6}
  \end{minipage}
\end{figure}

\paragraph{}

\begin{figure}[!h]
  \centering
  \begin{minipage}[b]{0.45\textwidth}
    \includegraphics[width=\textwidth]{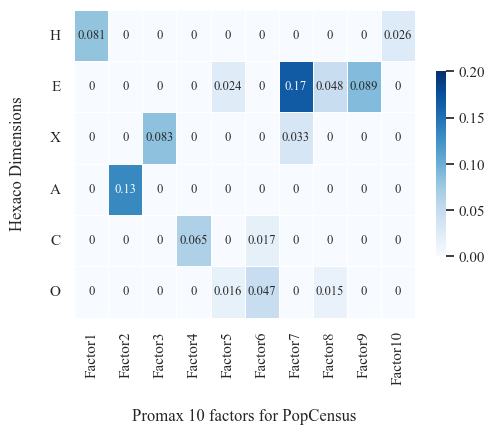}
    \caption{Weighted Jaccard similarity for Promax (10-Factor) results.}
    \label{fig:jaccard_10}
  \end{minipage}
  \hfill
  \begin{minipage}[b]{0.45\textwidth}
    \includegraphics[width=\textwidth]{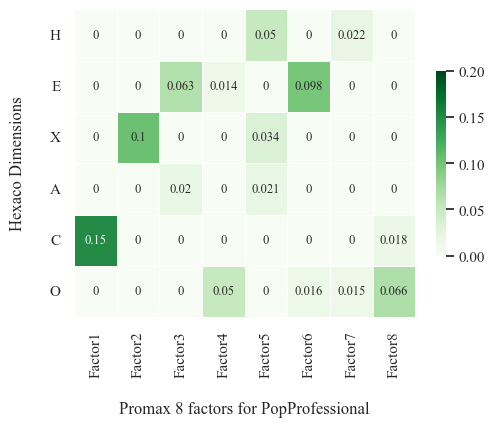}
    \caption{Weighted Jaccard similarity for Promax (8-Factor) results (PopProfessionals.}
    \label{fig:jaccard_8}
  \end{minipage}
\end{figure}

\newpage{}
\section{Appendix: Factor-6 results}
\label{app_6_results}

\begin{table}[!h]
    \centering
    \renewcommand{\arraystretch}{1}
    \setlength{\tabcolsep}{3pt}
    \begin{tabular}{lrlrlrlrlrlr}
    \hline
    \multicolumn{12}{c}{Factor} \\

    \multicolumn{2}{c}{(1) Dishonesty } 
    & \multicolumn{2}{c}{(2) Introversion } 
    & \multicolumn{2}{c}{(3) Unconventionality} 
    & \multicolumn{2}{c}{(4) Dominance}
    & \multicolumn{2}{c}{(5) Disagreeableness}
    & \multicolumn{2}{c}{(5) Provincial} \\
    
    \multicolumn{2}{c}{(4.24\% of variance)} 
    & \multicolumn{2}{c}{(3.96\% of variance)} 
    & \multicolumn{2}{c}{(3.79\% of variance)} 
    & \multicolumn{2}{c}{(3.06\% of variance)} 
    & \multicolumn{2}{c}{(2.70\% of variance)} 
    & \multicolumn{2}{c}{(1.96\% of variance)} \\
    
    Adjective & Load & Adjective & Load & Adjective & Load & Adjective & Load & Adjective & Load & Adjective & Load \\
    \hline
    
    sly &	0.91 &	ungregarious &	0.76 &	strait-laced &	-0.64 &	unbold &	0.72 &	sharp-tongued &	0.63 &	unliterary &	0.59 \\
    deceptive &	0.91 &	distant &	0.74 &	unconventional &	0.57 &	bullish &	-0.64 &	faultfinding &	0.60 &	unstudious &	0.57 \\
    manipulative &	0.90 &	untalkative &	0.73 &	conventional &	-0.57 &	mild &	0.64 &	overharsh &	0.58 &	unscholarly &	0.57 \\
    sneaky &	0.89 &	aloof &	0.72 &	hit-or-miss &	0.57 &	fierce &	-0.64 &	abrasive &	0.57 &	folksy &	0.56 \\
    devious &	0.88 &	detached &	0.72 &	overrigid &	-0.57 &	soft-spoken &	0.59 &	unforbearing &	0.56 &	homespun &	0.56 \\
    deceitful &	0.86 &	uncommunicative &	0.71 &	distractible &	0.56 &	placid &	0.58 &	peevish &	0.56 &	unbookish &	0.54 \\
    underhanded &	0.85 &	uncompanionable &	0.71 &	free-living &	0.55 &	forceful &	-0.58 &	argumentative &	0.56 &	sophisticated &	-0.54 \\
    double-faced &	0.85 &	taciturn &	0.70 &	unpredictable &	0.55 &	uncompetitive &	0.57 &	harsh &	0.56 &	rugged &	0.53 \\
    unscrupulous &	0.84 &	seclusive &	0.70 &	unreined &	0.54 &	sedate &	0.56 &	unforgiving &	0.55 &	overstudious &	-0.50 \\
    scheming &	0.84 &	undemonstrative &	0.70 &	impulsive &	0.54 &	forcible &	-0.55 &	caustic &	0.55 &	ultraintellectual &	-0.49 \\
    undeceptive &	-0.84 &	unaccessible &	0.68 &	spontaneous &	0.54 &	competitory &	-0.53 &	overbearing &	0.54 &	visionary &	-0.47 \\
    double-tongued &	0.84 &	withdrawn &	0.68 &	overscrupulous &	-0.54 &	self-assertive &	-0.53 &	reproachful &	0.53 &	cosmopolitan &	-0.47 \\
    undevious &	-0.83 &	unsociable &	0.67 &	unconstrained &	0.53 &	leisurely &	0.53 &	ultracritical &	0.53 &	old-fashioned &	0.46 \\
    pretenseful &	0.82 &	unsocial &	0.67 &	overconscientious &	-0.53 &	hectic &	-0.52 &	testy &	0.52 &	ultrarefined &	-0.46 \\
    incorrupt &	-0.81 &	asocial &	0.67 &	conservative &	-0.53 &	daring &	-0.52 &	overcritical &	0.52 &	unprovincial &	-0.46 \\
    untrustful &	0.81 &	unsolicitous &	0.67 &	messy &	0.53 &	venturesome &	-0.51 &	hypercritical &	0.52 &	cerebral &	-0.45 \\
    uncandid &	0.81 &	standoffish &	0.66 &	stringent &	-0.53 &	unaggressive &	0.51 &	nagging &	0.52 &	unprogressive &	0.45 \\
    untrustworthy &	0.80 &	impersonal &	0.65 &	unspontaneous &	-0.52 &	unhurried &	0.51 &	unaccommodating &	0.51 &	unskeptical &	0.44 \\
    evasive &	0.80 &	reclusive &	0.65 &	rigid &	-0.52 &	fiery &	-0.50 &	intolerant &	0.50 &	abstract &	-0.44 \\
    self-seeking &	0.79 &	hermitish &	0.65 &	footloose &	0.52 &	overdaring &	-0.50 &	censorial &	0.50 &	progressive &	-0.43 \\
    exploitative &	0.77 &	warmthless &	0.64 &	nonrigid &	0.51 &	lion-hearted &	-0.49 &	good-tempered &	-0.50 &	unsophisticated &	0.43 \\
    knavish &	0.77 &	sociable &	-0.64 &	devil-may-care &	0.51 &	sportsmanlike &	-0.48 &	crabby &	0.50 &	changeless &	0.42 \\
    dishonest &	0.77 &	unpersonable &	0.64 &	unbusinesslike &	0.51 &	unobtrusive &	0.48 &	forgiving &	-0.50 &	refined &	-0.42 \\
    untransparent &	0.77 &	social &	-0.64 &	strict &	-0.51 &	hot-blooded &	-0.48 &	forbearing &	-0.50 &	unchanging &	0.41 \\
    roguish &	0.77 &	withdrawing &	0.63 &	changeable &	0.51 &	quiet-spoken &	0.48 &	tolerant &	-0.49 &	conventional &	0.41 \\
    cagey &	0.76 &	dissocial &	0.63 &	unanchored &	0.50 &	audacious &	-0.48 &	undiplomatic &	0.49 &	urbane &	-0.41 \\
    plain-dealing &	-0.76 &	reserved &	0.63 &	undeviating &	-0.50 &	hard-nosed &	-0.48 &	nonirritable &	-0.48 &	unphilosophical &	0.41 \\
    unethical &	0.75 &	close-mouthed &	0.63 &	planless &	0.50 &	headstrong &	-0.48 &	unreasonable &	0.48 &	chic &	-0.41 \\
    incorruptible &	-0.74 &	antisocial &	0.61 &	orderly &	-0.50 &	unadventurous &	0.48 &	unagreeable &	0.48 &	bookish &	-0.40 \\
    ingratiatory &	0.74 &	unvocal &	0.61 &	overneat &	-0.50 &	pacifistic &	0.48 &	touchy &	0.48 &	brainy &	-0.40 \\
    \hline
    \end{tabular}
    \caption{Highest loaded terms recovered from the 6-Factor solution.}
    \label{tab:factor_6}
\end{table}

\newpage{}

\section{Appendix: Factor-10 results}
\label{app_10_results}

\begin{table}[!h]
    \centering
    \renewcommand{\arraystretch}{0.9}
    \setlength{\tabcolsep}{3pt}
    \begin{tabular}{lrlrlrlrlr}
    \hline
    \multicolumn{10}{c}{Factor} \\

    \multicolumn{2}{c}{(1) Dishonesty } 
    & \multicolumn{2}{c}{(2) Disagreeableness } 
    & \multicolumn{2}{c}{(3) Introversion} 
    & \multicolumn{2}{c}{(4) Unconscientiousness}
    & \multicolumn{2}{c}{(5) Unheroic} \\
    
    \multicolumn{2}{c}{(4.00\% of variance)} 
    & \multicolumn{2}{c}{(3.79\% of variance)} 
    & \multicolumn{2}{c}{(3.18\% of variance)} 
    & \multicolumn{2}{c}{(3.09\% of variance)} 
    & \multicolumn{2}{c}{(2.80\% of variance)} \\
    
    Adjective & Load & Adjective & Load & Adjective & Load & Adjective & Load & Adjective & Load \\
    \hline

    sly &	0.94 &	sharp-tongued &	0.74 &	uncommunicative &	0.72 &	overneat &	-0.67 &	unbold &	0.69 \\
    sneaky &	0.93 &	abrasive &	0.69 &	aloof &	0.71 &	overconscientious &	-0.60 &	unadventurous &	0.64 \\
    deceptive &	0.93 &	unforbearing &	0.67 &	untalkative &	0.71 &	messy &	0.60 &	venturesome &	-0.63 \\
    devious &	0.91 &	harsh &	0.66 &	seclusive &	0.69 &	hit-or-miss &	0.59 &	mild &	0.62 \\
    undevious &	-0.90 &	overharsh &	0.66 &	ungregarious &	0.68 &	unbusinesslike &	0.58 &	venturous &	-0.61 \\
    undeceptive &	-0.89 &	argumentative &	0.66 &	uncompanionable &	0.68 &	planless &	0.57 &	sedate &	0.61 \\
    manipulative &	0.89 &	unaccommodating &	0.65 &	distant &	0.67 &	unmethodical &	0.57 &	fierce &	-0.60 \\
    deceitful &	0.87 &	caustic &	0.65 &	detached &	0.67 &	unsystematic &	0.56 &	daring &	-0.60 \\
    underhanded &	0.86 &	overbearing &	0.65 &	withdrawn &	0.67 &	ultrafastidious &	-0.55 &	hectic &	-0.57 \\
    uncandid &	0.85 &	unforgiving &	0.64 &	taciturn &	0.67 &	overscrupulous &	-0.54 &	placid &	0.56 \\
    untrustful &	0.85 &	unconciliatory &	0.62 &	standoffish &	0.66 &	orderly &	-0.54 &	restless &	-0.56 \\
    double-faced &	0.84 &	faultfinding &	0.60 &	unsociable &	0.66 &	unreined &	0.53 &	unconstrainable &	-0.55 \\
    scheming &	0.84 &	testy &	0.60 &	unaccessible &	0.65 &	overrigorous &	-0.53 &	audacious &	-0.53 \\
    unscrupulous &	0.84 &	peevish &	0.59 &	asocial &	0.65 &	untidy &	0.53 &	adventurous &	-0.53 \\
    double-tongued &	0.84 &	brusque &	0.59 &	unsocial &	0.65 &	undisciplined &	0.53 &	fiery &	-0.53 \\
    untrustworthy &	0.82 &	good-tempered &	-0.59 &	withdrawing &	0.65 &	overdiligent &	-0.53 &	inexcitable &	0.52 \\
    untransparent &	0.82 &	unagreeable &	0.58 &	reclusive &	0.65 &	distractible &	0.52 &	irrestrainable &	-0.52 \\
    evasive &	0.82 &	forgiving &	-0.58 &	dissocial &	0.64 &	punctual &	-0.52 &	unventurous &	0.50 \\
    incorrupt &	-0.81 &	intolerant &	0.58 &	unsolicitous &	0.62 &	unheedful &	0.52 &	intense &	-0.50 \\
    cagey &	0.80 &	antagonistic &	0.58 &	undemonstrative &	0.62 &	stringent &	-0.52 &	untamable &	-0.49 \\
    roguish &	0.79 &	undiplomatic &	0.57 &	hermitish &	0.62 &	strait-laced &	-0.51 &	lion-hearted &	-0.49 \\
    pretenseful &	0.79 &	cantankerous &	0.57 &	impersonal &	0.62 &	heedless &	0.51 &	overdaring &	-0.49 \\
    plain-dealing &	-0.77 &	ungentle &	0.56 &	communicative &	-0.61 &	tidy &	-0.50 &	gutsy &	-0.49 \\
    suspicious &	0.76 &	reproachful &	0.56 &	close-mouthed &	0.60 &	overrigid &	-0.50 &	unrestrainable &	-0.49 \\
    exploitative &	0.75 &	forbearing &	-0.56 &	unpersonable &	0.60 &	conscientious &	-0.50 &	unheroic &	0.48 \\
    dishonest &	0.75 &	crabby &	0.56 &	antisocial &	0.59 &	unpunctual &	0.50 &	unexcitable &	0.48 \\
    knavish &	0.75 &	bossy &	0.56 &	social &	-0.59 &	lax &	0.50 &	hot-blooded &	-0.48 \\
    unethical &	0.73 &	high-handed &	0.55 &	unspeaking &	0.59 &	overfastidious &	-0.49 &	dynamic &	-0.48 \\
    incorruptible &	-0.73 &	censorial &	0.55 &	accessible &	-0.59 &	careless &	0.49 &	headlong &	-0.47 \\
    self-seeking &	0.72 &	tolerant &	-0.55 &	nonvocal &	0.59 &	perfectionistic &	-0.49 &	scrappy &	-0.47 \\
    \hline
    & & & & & & & & & \\
    \hline

    \multicolumn{10}{c}{Factor} \\

    \multicolumn{2}{c}{(6) Unscholarly } 
    & \multicolumn{2}{c}{(7) Gendered-emotionality } 
    & \multicolumn{2}{c}{(8) Unsentimentality} 
    & \multicolumn{2}{c}{(9) Insensitivity}
    & \multicolumn{2}{c}{(10) Unartistic} \\
    
    \multicolumn{2}{c}{(1.62\% of variance)} 
    & \multicolumn{2}{c}{(1.10\% of variance)} 
    & \multicolumn{2}{c}{(1.01\% of variance)} 
    & \multicolumn{2}{c}{(1.00\% of variance)} 
    & \multicolumn{2}{c}{(0.92\% of variance)} \\
    
    Adjective & Load & Adjective & Load & Adjective & Load & Adjective & Load & Adjective & Load \\
    \hline

    unbookish &	0.71 &	womanly &	1.07 &	unsentimental &	0.53 &	finicky &	-0.51 & ultrarefined &	-0.50 \\
    unscholarly &	0.70 &	gentlemanlike &	-0.96 &	earthy &	-0.49 &	fretful &	-0.48 &	chic &	-0.50 \\
    unliterary &	0.67 &	feminine &	0.90 &	homespun &	-0.49 &	fussy &	-0.45 &	lavish &	-0.49 \\
    bookish &	-0.62 &	masculine &	-0.89 &	long-suffering &	-0.39 &	overnervous &	-0.42 &	refined &	-0.45 \\
    ultraintellectual &	-0.61 &	manly &	-0.89 &	competitory &	0.38 &	worrying &	-0.41 &	dapper &	-0.45 \\
    overbookish &	-0.60 &	ladylike &	0.80 &	mechanistic &	0.38 &	anxious &	-0.37 &	elegant &	-0.45 \\
    scholarly &	-0.58 &	virile &	-0.77 &	sentimental &	-0.37 &	tense &	-0.36 &	overrefined &	-0.43 \\
    unphilosophical &	0.56 &	unmasculine &	0.58 &	tenderminded &	-0.37 &	overfastidious &	-0.36 &	unextravagant &	0.43 \\
    unskeptical &	0.55 &	rugged &	-0.43 &	careworn &	-0.37 &	perturbable &	-0.35 &	extravagant &	-0.41 \\
    overstudious &	-0.55 &	maternal &	0.43 &	tender-hearted &	-0.36 &	compulsive &	-0.35 &	cosmopolitan &	-0.41 \\
    literary &	-0.54 &	vivacious &	0.37 &	speedy &	0.36 &	supersensitive &	-0.34 &	polished &	-0.41 \\
    unstudious &	0.53 &	bubbly &	0.33 &	reverent &	-0.36 &	thick-skinned &	0.33 &	debonair &	-0.39 \\
    cerebral &	-0.51 &	sparkling &	0.31 &	serene &	-0.36 &	self-punishing &	-0.33 &	overelegant &	-0.38 \\
    inquisitorial &	-0.50 &	unbusinesslike &	0.31 &	folksy &	-0.35 &	self-reproachful &	-0.33 &	sophisticated &	-0.38 \\
    homespun &	0.49 &	perky &	0.30 &	tender &	-0.34 &	ultrafastidious &	-0.33 &	exclusive &	-0.38 \\
    philosophizing &	-0.48 &	graceful &	0.30 &	spiritual &	-0.34 &	imperturbable &	0.31 &	urbane &	-0.37 \\
    rugged &	0.47 &	sassy &	0.30 &	poetic &	-0.33 &	high-strung &	-0.31 &	genteel &	-0.34 \\
    studious &	-0.47 &	voluptuous &	0.29 &	overcaring &	-0.32 &	self-critical &	-0.31 &	untheatrical &	0.34 \\
    well-read &	-0.46 &	hardened &	-0.29 &	sensitive &	-0.32 &	touchy &	-0.31 &	inelegant &	0.34 \\
    folksy &	0.46 &	unfeminine &	-0.28 &	nonspiritual &	0.32 &	overneat &	-0.31 &	designful &	-0.34 \\
    brainy &	-0.45 &	emotional &	0.28 &	gentle-minded &	-0.32 &	faultfinding &	-0.31 &	fanciful &	-0.34 \\
    sophisticated &	-0.44 &	demure &	0.28 &	maternal &	-0.32 &	artistic &	-0.29 &	suave &	-0.33 \\
    unprovincial &	-0.43 &	angelic &	0.28 &	unartistic &	0.32 &	thin-skinned &	-0.29 &	overneat &	-0.30 \\
    encyclopedic &	-0.41 &	chivalrous &	-0.28 &	old-fashioned &	-0.30 &	overintense &	-0.29 &	snoopy &	0.30 \\
    abstract &	-0.41 &	elegant &	0.28 &	soft-hearted &	-0.30 &	overcurious &	-0.29 &	showy &	-0.28 \\
    arbitrative &	-0.39 &	chic &	0.28 &	ultrasentimental &	-0.30 &	hypersensitive &	-0.29 &	fussy &	-0.28 \\
    philosophical &	-0.39 &	dainty &	0.27 &	altruistic &	-0.29 &	unartistic &	0.29 &	unostentatious &	0.28 \\
    eloquent &	-0.37 &	lamblike &	0.27 &	feelingful &	-0.28 &	oversensitive &	-0.28 &	unartistic &	0.28 \\
    political &	-0.37 &	nonspiritual &	-0.27 &	dispassionate &	0.28 &	overscrupulous &	-0.28 &	artistic &	-0.28 \\
    earthy &	0.37 &	girlish &	0.26 &	humanitarian &	-0.28 &	overthoughtful &	-0.28 &	uncourtly &	0.27 \\
    \hline
    \end{tabular}
    \caption{Highest loaded terms recovered from the 10-Factor solution.}
    \label{tab:factor_10}
\end{table}

\newpage{}

\section{Appendix: Factor-8 results (PopProfessional)}
\label{app_8_results}

\begin{table}[!h]
    \centering
    \renewcommand{\arraystretch}{0.9}
    \setlength{\tabcolsep}{3pt}
    \begin{tabular}{lrlrlrlrlr}
    \hline
    \multicolumn{10}{c}{Factor} \\

    \multicolumn{2}{c}{(1) Unconscientiousness } 
    & \multicolumn{2}{c}{(2) Introversion } 
    & \multicolumn{2}{c}{(3) Unconciliatory} 
    & \multicolumn{2}{c}{(4) Conventional}
    & \multicolumn{2}{c}{(5) Extravagant } \\
    
    \multicolumn{2}{c}{(3.22\% of variance)} 
    & \multicolumn{2}{c}{(2.54\% of variance)} 
    & \multicolumn{2}{c}{(2.48\% of variance)} 
    & \multicolumn{2}{c}{(2.29\% of variance)} 
    & \multicolumn{2}{c}{(1.90\% of variance)} \\
    
    Adjective & Load & Adjective & Load & Adjective & Load & Adjective & Load & Adjective & Load \\
    \hline
    
    unmethodical &	0.74 &	undemonstrative &	0.60 &	brusque &	0.58 &	conventional &	0.68 &	unextravagant &	-0.55 \\
	unsystematic &	0.73 &	unvocal &	0.60 &	womanly &	-0.58 &	traditional &	0.60 &	acquisitive &	0.53 \\
	haphazard &	0.73 &	untalkative &	0.58 &	bullheaded &	0.56 &	conservative &	0.60 &	lavish &	0.53 \\
	planless &	0.70 &	ungregarious &	0.58 &	unconciliatory &	0.56 &	sedate &	0.59 &	unostentatious &	-0.52 \\
	scatterbrained &	0.66 &	unexpansive &	0.58 &	hard-nosed &	0.55 &	changeless &	0.57 &	untheatrical &	-0.50 \\
	overconscientious &	-0.66 &	seclusive &	0.57 &	blunt &	0.55 &	old-fashioned &	0.56 &	opportunistic &	0.50 \\
	unpunctual &	0.66 &	taciturn &	0.57 &	hardheaded &	0.52 &	unchanging &	0.55 &	extravagant &	0.49 \\
	careless &	0.65 &	nonvocal &	0.54 &	sharp-tongued &	0.51 &	unconventional &	-0.55 &	exclusive &	0.47 \\
	overdiligent &	-0.65 &	uncommunicative &	0.54 &	unforgiving &	0.50 &	nonvariant &	0.54 &	quiet-spoken &	-0.46 \\
	hit-or-miss &	0.65 &	hermitish &	0.54 &	harsh &	0.49 &	predictable &	0.52 &	showy &	0.46 \\
	disorganized &	0.64 &	reclusive &	0.54 &	overharsh &	0.49 &	strait-laced &	0.52 &	unassuming &	-0.46 \\
	heedless &	0.63 &	asocial &	0.53 &	unaccommodating &	0.49 &	unadventurous &	0.52 &	uncunning &	-0.46 \\
	undisciplined &	0.62 &	distant &	0.53 &	unbending &	0.49 &	inexcitable &	0.51 &	unbusinesslike &	-0.45 \\
	unthorough &	0.61 &	withdrawn &	0.53 &	pigheaded &	0.49 &	unvarying &	0.50 &	unpresuming &	-0.45 \\
	undeliberative &	0.61 &	detached &	0.53 &	obstinate &	0.49 &	unconstrainable &	-0.50 &	unexplosive &	-0.45 \\
	undiligent &	0.60 &	unconfiding &	0.52 &	stern &	0.48 &	restless &	-0.49 &	extroverted &	0.45 \\
	lax &	0.60 &	unpersonable &	0.52 &	ladylike &	-0.48 &	variable &	-0.49 &	rivalrous &	0.45 \\
	inconsistent &	0.60 &	close-mouthed &	0.52 &	testy &	0.48 &	unexcitable &	0.48 &	ingratiatory &	0.44 \\
	digressive &	0.60 &	unsocial &	0.52 &	uncomplaisant &	0.48 &	venturesome &	-0.47 &	modest &	-0.44 \\
	absent-minded &	0.60 &	reserved &	0.51 &	abrasive &	0.47 &	unprogressive &	0.47 &	earthy &	-0.44 \\
	planful &	-0.59 &	uncompanionable &	0.51 &	manly &	0.47 &	unchangeable &	0.46 &	dapper &	0.44 \\
	unexacting &	0.58 &	unsociable &	0.51 &	curt &	0.47 &	inconfinable &	-0.45 &	suave &	0.43 \\
	unvigilant &	0.57 &	withdrawing &	0.51 &	indocile &	0.46 &	unventurous &	0.45 &	unhurried &	-0.43 \\
	organized &	-0.57 &	dissocial &	0.51 &	masculine &	0.46 &	unadaptable &	0.44 &	debonair &	0.43 \\
	distractible &	0.56 &	unspeaking &	0.51 &	hardened &	0.46 &	ultraconservative &	0.44 &	chitchatty &	0.43 \\
	messy &	0.56 &	aloof &	0.51 &	stubborn &	0.46 &	venturous &	-0.43 &	slick &	0.42 \\
	orderly &	-0.56 &	unsolicitous &	0.50 &	inflexible &	0.45 &	unmodifiable &	0.43 &	rugged &	-0.42 \\
	unheedful &	0.56 &	impersonal &	0.49 &	unagreeable &	0.45 &	uneager &	0.43 &	bullish &	0.42 \\
	punctual &	-0.55 &	uncourtly &	0.49 &	forgiving &	-0.45 &	invariable &	0.42 &	down-to-earth &	-0.42 \\
	meticulous &	-0.55 &	standoffish &	0.48 &	ultracritical &	0.45 &	unstirrable &	0.42 &	chic &	0.41 \\
    \hline
    
    & & & & & & & & & \\
    \cline{1-6}

    \multicolumn{6}{c}{Factor} \\

    \multicolumn{2}{c}{(6) Emotionality } 
    & \multicolumn{2}{c}{(7) Heroic } 
    & \multicolumn{2}{c}{(8) Unscholarly } 
    & \multicolumn{2}{c}{ }
    & \multicolumn{2}{c}{ } \\
    
    \multicolumn{2}{c}{(1.51\% of variance)} 
    & \multicolumn{2}{c}{(1.22\% of variance)} 
    & \multicolumn{2}{c}{(0.89\% of variance)} 
    & \multicolumn{2}{c}{ } 
    & \multicolumn{2}{c}{ } \\
    
    Adjective & Load & Adjective & Load & Adjective & Load & & & & \\
    \cline{1-6}

    unsentimental &	-0.58 &	finicky &	-0.59 &	bookish &	-0.66 \\
    unbusinesslike &	0.52 &	artistic &	-0.52 &	unbookish &	0.65 \\
    calculating &	-0.52 &	fanciful &	-0.50 &	unphilosophical &	0.56 \\
    supersensitive &	0.48 &	fussy &	-0.49 &	literary &	-0.55 \\
    ultrasentimental &	0.47 &	chic &	-0.48 &	unscholarly &	0.53 \\
    overemotional &	0.46 &	overfastidious &	-0.48 &	overbookish &	-0.50 \\
    fretful &	0.44 &	ultrarefined &	-0.43 &	scholarly &	-0.50 \\
    sentimental &	0.44 &	heroic &	0.43 &	overstudious &	-0.47 \\
    thin-skinned &	0.44 &	designful &	-0.42 &	philosophizing &	-0.44 \\
    thick-skinned &	-0.44 &	overneat &	-0.42 &	unliterary &	0.44 \\
    businesslike &	-0.44 &	picky &	-0.40 &	ultraintellectual &	-0.44 \\
    hypersensitive &	0.43 &	crusading &	0.39 &	studious &	-0.43 \\
    uncalculating &	0.43 &	cosmopolitan &	-0.39 &	encyclopedic &	-0.42 \\
    soft-shelled &	0.43 &	poetic &	-0.38 &	wordy &	-0.41 \\
    tenderminded &	0.42 &	elegant &	-0.38 &	well-read &	-0.40 \\
    hard-nosed &	-0.41 &	discriminative &	-0.38 &	philosophical &	-0.39 \\
    melancholic &	0.40 &	abstract &	-0.37 &	inquisitorial &	-0.38 \\
    mechanistic &	-0.40 &	overparticular &	-0.37 &	unpolitical &	0.38 \\
    overcaring &	0.39 &	valiant &	0.37 &	long-winded &	-0.30 \\
    self-conscious &	0.38 &	valorous &	0.37 &	unskeptical &	0.30 \\
    competitory &	-0.38 &	overrefined &	-0.36 &	cerebral &	-0.30 \\
    tender &	0.38 &	ultrafastidious &	-0.36 &	unstudious &	0.29 \\
    oversensitive &	0.38 &	refined &	-0.36 &	unquestioning &	0.27 \\
    touchy &	0.38 &	overimaginative &	-0.35 &	careworn &	-0.27 \\
    earthy &	0.37 &	unartistic &	0.35 &	overcurious &	-0.27 \\
    calculable &	-0.37 &	uncontriving &	0.34 &	ineloquent &	0.26 \\
    poetic &	0.37 &	urbane &	-0.34 &	nosy &	-0.26 \\
    nonspiritual &	-0.36 &	unpleasable &	-0.34 &	eloquent &	-0.26 \\
    tough-minded &	-0.36 &	scrappy &	0.34 &	verbal &	-0.26 \\
    \cline{1-6}
    
    \end{tabular}
    \caption{Highest loaded terms recovered from the 8-Factor solution, using PopProfessional.}
    \label{tab:factor_8}
\end{table}

\end{document}